\documentclass[manuscript,screen]{acmart}
\usepackage{multirow}

\usepackage{algorithmic}
\usepackage{graphicx}
\usepackage{textcomp}
\usepackage{xcolor}
\usepackage{hyperref}

\AtBeginDocument{%
  \providecommand\BibTeX{{%
    \normalfont B\kern-0.5em{\scshape i\kern-0.25em b}\kern-0.8em\TeX}}}

\setcopyright{acmcopyright}
\copyrightyear{2023}
\acmYear{2023}
\acmDOI{XXXXXXX.XXXXXXX}

\acmJournal{JACM}
\acmVolume{37}
\acmNumber{4}
\acmArticle{111}
\acmMonth{8}

\acmPrice{15.00}
\acmISBN{978-1-4503-XXXX-X/18/06}

\begin{document}

\title{3D Face Reconstruction: the Road to Forensics}

\author{Simone Maurizio La Cava}
\email{simonem.lac@unica.it}
\orcid{0000-0002-6344-1845}
\affiliation{%
  \institution{University of Cagliari}
  \city{Cagliari}
  \country{Italy}
}

\author{Giulia Orrù}
\email{giulia.orru@unica.it}
\orcid{0000-0002-7802-2483}
\affiliation{%
  \institution{University of Cagliari}
  \city{Cagliari}
  \country{Italy}
}

\author{Martin Drahansky}
\email{martin.drahansky@tbs-biometrics.com}
\orcid{0000-0002-9321-7385}
\affiliation{%
    \institution{Touchless Biometric Systems (TBS) Holding AG} \country{Switzerland}
}

\author{Gian Luca Marcialis}
\email{marcialis@unica.it}
\orcid{0000-0002-8719-9643}
\affiliation{%
  \institution{University of Cagliari}
  \city{Cagliari}
  \country{Italy}
}

\author{Fabio Roli}
\email{fabio.roli@unige.it}
\orcid{0000-0003-4103-9190}
\affiliation{%
  \institution{University of Genova}
  \city{Genova}
  \country{Italy}
}

\renewcommand{\shortauthors}{La Cava et al.}

\begin{abstract}
3D face reconstruction algorithms from images and videos are applied to many fields, from plastic surgery to the entertainment sector, thanks to their advantageous features. However, when looking at forensic applications, 3D face reconstruction must observe strict requirements that still make its possible role in bringing evidence to a lawsuit unclear. An extensive investigation of the constraints, potential, and limits of its application in forensics is still missing. Shedding some light on this matter is the goal of the present survey, which starts by clarifying the relation between forensic applications and biometrics, with a focus on face recognition. Therefore, it provides an analysis of the achievements of 3D face reconstruction algorithms from surveillance videos and mugshot images and discusses the current obstacles that separate 3D face reconstruction from an active role in forensic applications. Finally, it examines the underlying data sets, with their advantages and limitations, while proposing alternatives that could substitute or complement them.
\end{abstract}

\ccsdesc[500]{Applied computing~Investigation techniques}
\ccsdesc[500]{Social and professional topics~Surveillance}
\ccsdesc[500]{Computing methodologies~Biometrics}
\ccsdesc[500]{Computing methodologies~3D imaging}
\ccsdesc[500]{Computing methodologies~Reconstruction}
\ccsdesc[500]{Computing methodologies~Shape inference}
\ccsdesc[300]{Computing methodologies~Shape representations}
\ccsdesc[300]{Computing methodologies~Appearance and texture representations}

\keywords{3D face reconstruction, forensics, recognition}

\maketitle

\section{Introduction}
In the last few decades, much attention has been paid to the use of 3D data in facial image processing applications. This technology has shown to be promising for robust facial feature extraction \cite{2D3Dsurvey, robust, robustsurvey}. In uncontrolled environments, it limits the effects of adverse factors such as unfavourable illumination conditions and the non-frontal poses of the face with respect to the camera \cite{robust,uncalibrated, Carabinieri}. 

\begin{figure*}[]
	\centering
	\includegraphics[width=0.8\linewidth]{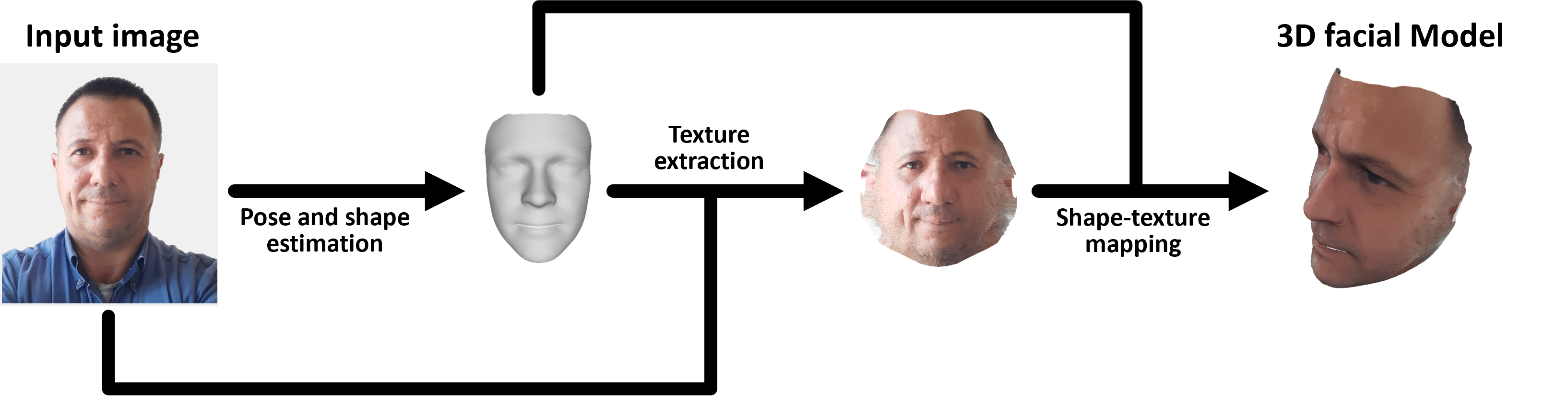}
	\caption{\textit{Example of reconstruction of a 3D facial model from a single input 2D image (obtained through the framework proposed by \cite{eos}).}}
	\label{fig:3DFR}
\end{figure*}

Among the various scenarios, developing personal recognition based on 3D data appears to be a "hot topic" due to the accuracy and efficiency obtainable from comparing faces, thanks to the complementary information of shape and texture  \cite{Afzal2020, 3Dstereo, ICPrecognition}. However, acquiring such data requires expensive hardware; moreover, the enrolment process is much more complex \cite{dis3D, dis3D2, dis3D3, uncalibrated, dis3D4}. Thus, face recognition technology was mainly developed in the 2D domain. The acquisition of 2D images is more straightforward than that of 3D ones, as it does not require specific hardware, but often makes the recognition task challenging due to the significant variability in facial appearance \cite{Cadoni2010, uncalibrated}. 3D face reconstruction (3DFR) from 2D images and videos may overcome these limits, combining the ease of acquiring 2D data with the robustness of 3D ones (Fig. \ref{fig:3DFR}).

\begin{figure*}[!ht]
	\centering
	\includegraphics[width=0.75\linewidth]{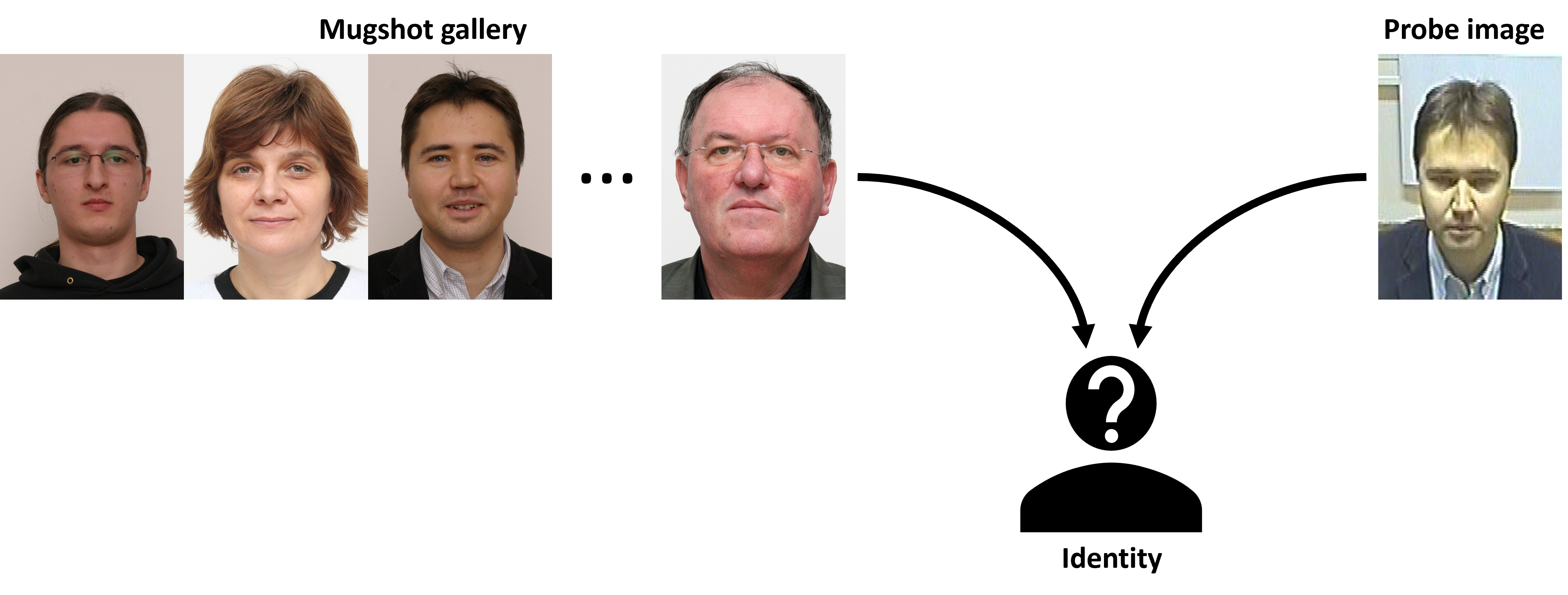}
	\caption{\textit{Example of forensic facial recognition from a mugshot reference gallery and a probe image (images from the SCface dataset \cite{SCFace}).}}
	\label{fig:mugshotvsprobe}
\end{figure*}

One of the possible fields that could benefit from these advantageous characteristics is that of forensics, which often deals with probe images of unidentified people's faces in non-frontal view, in uncontrolled environments, and in an uncooperative way, such as in the case of the ones captured by CCTV (Closed-Circuit Television) cameras. Despite some frameworks for the acquisition of 3D face models of suspects have been proposed (\textit{e.g.}, \cite{3Dmugshot}), in such context, it is still common to have 2D mugshots, that is, frontal and, usually, profile images of subjects routinely captured by law enforcement agencies \cite{Liang2018} for the recognition of people of interest, such as suspects or witnesses (Fig. \ref{fig:mugshotvsprobe}). 

Unfortunately, a reference gallery composed of frontal and profile images is not able to provide effective coverage of all possible conditions, such as in the case of a probe image in an arbitrary pose which is not at the same view angle as in one of the available mugshot images \cite{Zhang2008}. Therefore, from the first attempt at face recognition from mugshots \cite{firstMugRec}, 3D reconstruction techniques were exploited too for facing some of the issues which are typical of the considered forensic cases, trying to establish the identity of unknown individuals against a reference data set of known individuals, either in verification mode (1 to 1) or identification mode (1 to N). Hence, the research community proposed to employ this approach in facial recognition from probe videos and images acquired in an unconstrained environment to provide more information about the individual faces through the generation of multiple views or the "correction" of the pose in probe data. This makes the comparison with reference data more robust to various appearance variations typical of forensic cases.

In particular, to be suitable for real-world forensic applications, any system of this kind should satisfy strict constraints leading to the legal validity of the conclusions during a lawsuit or in the investigation phase \cite{JainSurvey, forensicRequirements}. For this reason, it is necessary to analyze the methods which employ 3DFR to shed some light on their admissibility in the forensic scenario. Although other authors investigated the state of the art of 3DFR from 2D images or videos \cite{monocular2018, monocular2021, statistical2020, uncalibrated} and its applications to face recognition \cite{frontal2020, statistical2020, uncalibrated}, none of them considered the requirements they have to satisfy to be potentially employed in such context and how forensics can benefit from their adoption. 
Moreover, the validity of the proposed face recognition systems in the considered application scenarios strongly depends on the data sets on which they have been evaluated since these provide a basis for measuring and comparing their performance with state of the art. In other words, data representativeness is fundamental, and the algorithms' adoption is bounded by the available data \cite{2DDatabasesSurvey, databasesSurvey}. 

A specific investigation highlighting the potential and limits of 3D facial reconstruction in forensics is still missing, and, in our opinion, it would be necessary to direct research toward its real-world application.
In order to pursue this goal, this work analyzes the potentiality of the employment of 3D face reconstruction in forensics and the approaches proposed by the research community for its integration in a common face recognition casework while considering the core challenges of legal admissibility of automated systems including it. The central premise of this work is to shed some light on the requirements that should be satisfied to fill the gap between biometric recognition and forensic comparison when reconstructing a facial image into 3D space for the recognition of an individual from 2D videos or images. The investigation of the potential benefit of this technique to forensics is the aim of our work. 

\begin{figure*}[]
	\centering
	\includegraphics[width=1\linewidth]{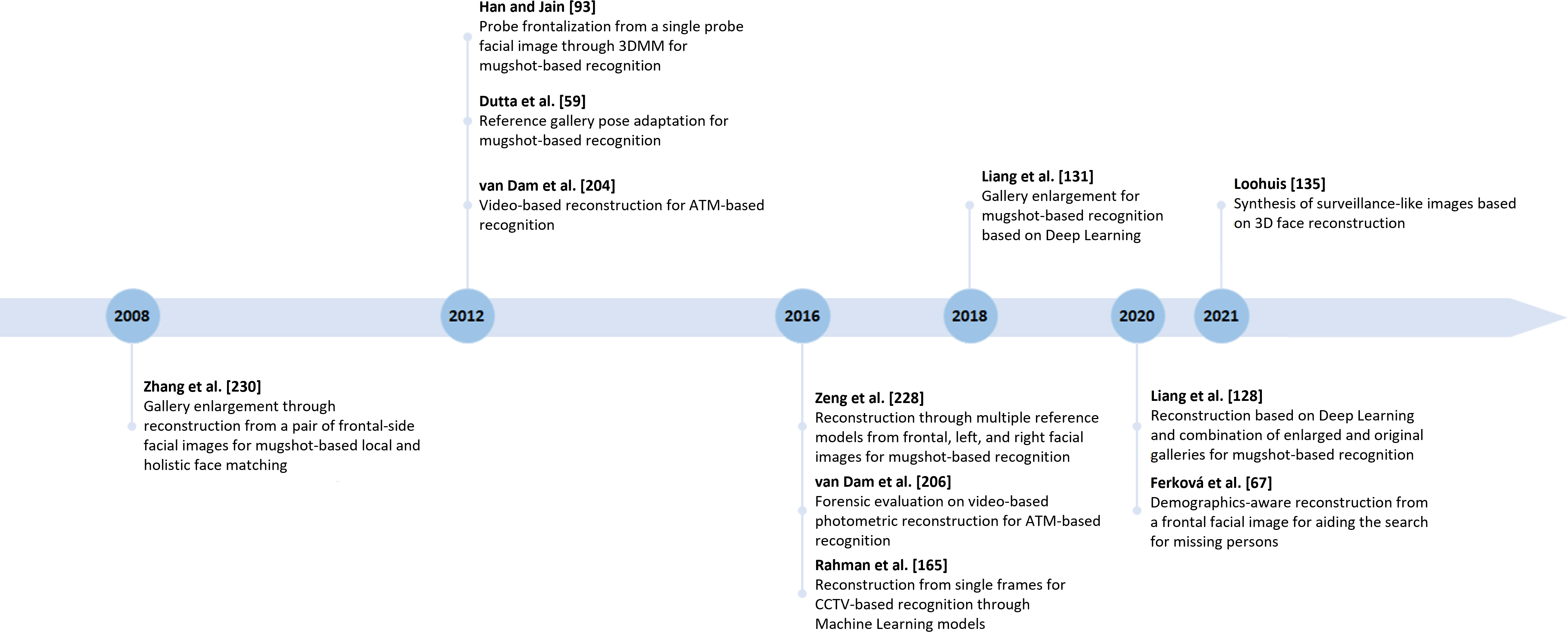}
	\caption{\textit{Milestones of forensic identification based on 3D face reconstruction.}}
	\label{fig:milestones}
\end{figure*}

This paper is the follow-up of Ref. \cite{la20223d}, which is a first step toward the objectives listed above. To our knowledge, it represents the first investigation focused on state of the art in applications and potentialities of 3D face reconstruction in forensics and the novelties introduced to date (Fig. \ref{fig:milestones}), as well as the requirements that any of the related systems must satisfy to be considered admissible in criminal investigations or judicial cases.
With respect to \cite{la20223d}, this paper extends such disquisition, especially in relation to the comparison among the proposed methods and the admissibility constraints which have to be satisfied in order to be effectively integrated into the reference scenario.
Moreover, this survey also provides an analysis of the data sets employed in the reviewed studies, which could further highlight their strengths and limits, suggesting their uses in the design and evaluation of forensic facial recognition algorithms and the potential issues. Finally, some state-of-the-art data sets which could be alternative or complementary to those already used are proposed and analyzed as well to provide suitable ground truth for future studies, with the main focus on the types of data so far considered, namely facial images, videos and 3D scans of the face.

The paper's structure is as follows. 
Section \ref{sec:biometrics} analyzes the relationship between forensics and biometrics, mainly focusing on facial traits and the integration of 3DFR. 
The state-of-the-art assessment of 3DFR methods for face recognition from mugshot images is reported in Section \ref{sec:mugshot}. 
A review of other proposed forensic-related applications of 3DFR from facial images and videos is carried out in Section \ref{sec:other}. 
Section \ref{sec:databases} explores the underlying data sets of facial images, videos, and 3D scans, proposing others which could be suitable as well for future research on the analyzed topic.
Finally, Section \ref{sec:conclusion} discusses how all the aspects above converge in a unified view.

\section{Face recognition and forensics}\label{sec:biometrics}

The face represents a valuable clue in many criminal investigations due to its advantageous characteristics with respect to other biometrics \cite{recognitionsurvey, faceInvestigation} and the growing number of surveillance cameras in both private and public places \cite{surveillanceMarket, DeLaTorre2015, Hu2017}. Over the years, various methods have been proposed to check whether the individual's identity in a probe image or video matches that of a person of interest, namely an individual related to the event under investigation, such as a suspect, a victim, or a witness. In particular, these represent a subset of the approaches widely explored in traditional biometric recognition and implemented in the related automated face recognition systems \cite{holisticRecognition, photogrammetryRecognition, recognitionsurvey}.
These methods can be summarized into various qualitative or quantitative examination approaches, which can be employed or are preferred under different conditions \cite{Edmond2009, ENFSIcomparison}.

A first approach processes the face globally in a holistic form.
However, it is only recommended where other more effective are not suitable and is highly inaccurate when faces belong to unfamiliar people, in the case of partially occluded faces \cite{ENFSIcomparison, unfamiliar, Zeinstra2018, humanFactor, Estudillo2021} or severely distorted CCTV footage \cite{Burton1999}.

A second approach is based on a set of facial fiducial points named landmarks \cite{landmarks, landmarkclasses} and employed to derive the distances and proportions between facial features. This choice is not generally recommended as well due to the subjectivity in their manual estimation in uncontrolled images due to adverse factors such as the large pose of the head, the distance from the camera, facial expressions, and lighting conditions \cite{ENFSIcomparison, landmarkUnreliability, landmarkUnreliability2, Moreton2021, VeraRodriguez2013}. Some of these issues could be mitigated by means of preprocessing techniques (\textit{e.g.}, super-resolution methods \cite{hong2022facial}).

A third approach is that of superimposition. It allows handling the discrepancies arising from differences in the position of the face with respect to the camera in two different aligned images or videos. To achieve this goal, it combines them through various methods, such as a reduced opacity overlay or blinking quickly between them. 
This approach is unreliable when comparing data acquired in uncontrolled scenarios, even in previous judicial cases \cite{superimposition, ENFSIcomparison, chimeric, superimpositionFail, Zeinstra2018, MalletEvison2013, Moreton2021, RvClarke}.

A fourth approach is that of morphological comparison, in which a generally predefined list of facial regions and features extracted from them related to shape, appearance, presence and/or location, such as the relative width of the mouth with respect to the distance between the eyes and the asymmetry of the mouth \cite{FISWGmorph}, are compared to determine differences and similarities between the probe and reference data \cite{Zeinstra2018}. 

In particular, the latter approach is able to improve the identification accuracy by examiners, even thanks to the higher physical stability over time with respect to many of both photoanthropometry and holistic features \cite{FISWGstability, landmarkUnreliability}. However, the stability of the evaluated features could also be affected by extrinsic factors, such as lighting and the position of the subject's face with respect to the camera, which can introduce different levels of variability, contributing to the unreliability of certain features \cite{landmarkUnreliability, landmarkFailure, landmarkFailure3D}.

Despite the differences in reliability and acceptance, these approaches are not alternatives to each other. The choice among them is generally dependent on the probe image or videos, and they can even be used jointly in the identification task to carry out a more exhaustive analysis \cite{faceforensics, automatedFaces, landmarkUnreliability}. Furthermore, even if these approaches could not be used as evidence in a confirmatory identification due to the acquisition condition of the probe image or video, these could still be employed in an attempt to exclude possible suspects or be a limited - but not worthless - support for reaching a conclusion through other evidence \cite{FISWGoverview, landmarkUnreliability, landmarkFailure, MalletEvison2013}.

Although both biometric recognition and forensic identification seek to link evidence to a particular individual \cite{JainRoss2015}, research in these fields has been pursued independently for many years due to their different goals and requirements, as well as the difficulties in achieving significant scientific contributions in this cross-domain research field \cite{la20223d}. Thus, despite the employment of approaches which are common between them, the underlying methods and the automated systems integrating them must satisfy strict constraints in order to be considered suitable for forensic casework.

\subsection{Automated Forensic Facial Recognition: The Italian Case} \label{subsec:italian}

Due to the stringent requirements of the analyzed field, automatic recognition systems are only recently being introduced. 
For example, in 2017, the Italian police bodies introduced the ordinary use of an automatic image recognition system, S.A.R.I. (from the Italian "Sistema Automatico di Riconoscimento delle Immagini"), as an innovative tool aimed at supporting investigative activities \cite{italian, ANSA}.
This system allows automatically comparing a facial probe image with millions of mugshots to reduce the number of candidates, which are then ordered by the similarity degree. Furthermore, the system is also able to work in real-time on a gallery on the order of hundreds of thousands of individuals to enforce security and control on the territory.
The SARI's outcome is a set of potential candidates that must be examined by the specialized experts of the scientific police in charge of verifying the process \cite{italian,barroccu2013prova}.
Due to the stringent requirements of the analyzed field, automatic recognition systems are only recently being exploited. 
Despite the effectiveness and the extreme speed of this automatic system, it cannot yet be used in the criminal field, as it does not allow access and repetition of recognition by the defense, thus precluding cross-examination of the specific functioning of the software in question \cite{italian, Scalfati, CarlizziTuzet, Ferrua}. Moreover, its functioning lacks the transparency required for any criminal case, thus precluding its compatibility with the constitutional procedural guarantees granted to the suspect \cite{italian, Scalfati}. 

If this is the state of things in face recognition, what about 3D face reconstruction? 3D reconstruction is already employed for enhancing the views of crime scenes (\textit{e.g.}, \cite{Meredith}), as computer-generated evidence \cite{barroccu2013prova}. Thanks to the 3D representation of the scene, obtained by one or more reference photographs, it is possible to recreate the aimed scenario, for example, by inserting moving objects and simulating people's behaviour while respecting the physical laws. 
However, depending on the task for which it has to be employed, this technology could be considered not admissible due to the still experimental nature of the underlying method \cite{barroccu2013prova}. Furthermore, the accuracy of the reconstruction of the human body is low, and the face is strongly influenced by the definition of the reference images as well as by the subjectivity of the operator in positioning the characterizing points for the reconstruction \cite{MastronardiVilardi}.

Therefore, a fully automated 3D reconstruction such as the one integrated into biometric systems could reduce the errors caused by the operator, standardize the process, and speed up the analysis, provided that sufficient quality of the resulting 3D model can be guaranteed. These advantages led the research community to propose methods and approaches strictly focused on reconstructing the body or even single parts, such as the face. In particular, the 3D reconstruction of the face could be crucial for some forensic recognition tasks, strongly enhancing the recognition accuracy with respect to the recognition from raw images, especially on faces represented in non-frontal poses. In particular, this technology could even be integrated into the previously cited scene reconstruction technology to enforce the reliability of the related computer-generated evidence and make it employable for real recognition tasks. 

\begin{figure*}[h]
	\centering
	\includegraphics[width=0.8\linewidth]{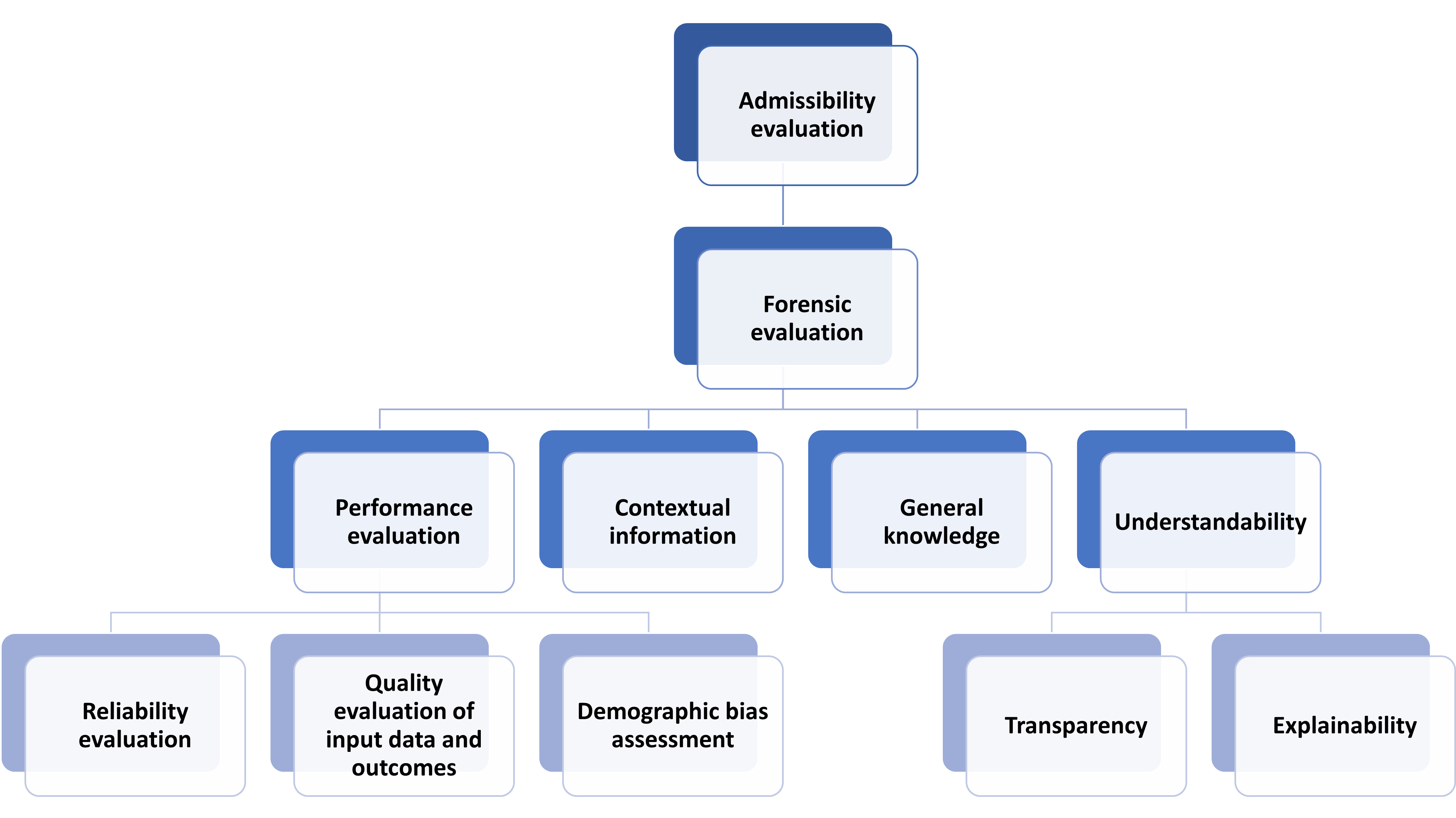}
	\caption{\textit{Forensic admissibility evaluation of an automatic biometric system in a casework.}}
	\label{fig:admissibility}
\end{figure*}

These factors could be crucial in the introduction of this technology in the forensic recognition task. However, it must comply with the technical and admissibility requirements, summarized in Figure \ref{fig:admissibility} and discussed in the following subsections, which any system must satisfy to be considered suitable to be employed in such a field.

\subsection{Biometric Systems and Forensic Admissibility} \label{subsec:admissibility}

Techniques and systems designed for biometrics, especially the automated ones, are appealing for their potential to address some forensic domain’s problems concerning crime prevention, crime investigation, and judicial trials in a more efficient, "scientifically objective", and standardized way \cite{YanOsadciw2008, JainRoss2015, faceforensics, forensicExaminers, localmarks, ethicsPolicy, Carabinieri}. In the case of face recognition, the related recognition technology has a role in many forensic and security applications, such as in identifying people of interest (\textit{e.g.}, terrorists) and searching for missing people, even in real-time \cite{Galterio2018, Hill2022, TheGuardianBeer}. In particular, concepts behind biometric facial recognition could be beneficial in various tasks underlying forensic applications. For example, person re-identification and face identification could aid the search task of forensic practitioners, thus the collection of evidence from crime scene images acquired from surveillance cameras \cite{Tistarelli2014}, and the investigation, thus linking traces between crime scenes by generating and testing likely explanations \cite{Tistarelli2014}. Face recognition could as well represent an aid in the individualization (or forensic evaluation) step, in which the evidential value is computed and assigned to the collected traces \cite{Tistarelli2014}, with noticeable parallelism with the similarity scores assigned by most automated face recognition systems in biometric recognition tasks.

However, despite several groups, such as the FISWG (Facial Identification Scientific Working Group) \cite{FISWGweb} and the ENFSI (European Network of Forensic Science Institutes) \cite{ENFSIweb}, which are currently working in this direction, there is no standardized and validated method in forensics \cite{faceforensics, guidelines, ethicsPolicy}. 
For example, in the United States, the admissibility of scientific evidence obtained through face recognition is generally evaluated through two guidelines: 

\begin{itemize}
    \item the ``Frye's rule'' gives the judges the task of assessing whether the technique or technology is accepted in a relevant scientific community \cite{FryeCase};
    \item the ``Daubert's rule'' adds to the previous one the constraints that it has been tested, the description of its error rate is available, and it must be maintained and adhere to standards \cite{faceforensics, DaubertCase, Erickson, DaubertCase2, DaubertCase3}.
\end{itemize}

In many other judicial systems beyond the U.S.A., no specific admissibility rule regarding the evaluation of the scientific evidence is given, such as the case of the European judicial system, where the judges are generally responsible for its assessment in single cases \cite{faceforensics}.
Another issue is the general acceptance of the biometric itself, especially the face, to the point that some governments banned or limited its usage even in law enforcement agencies (\textit{e.g.}, \cite{CNNBan, NewYorkTimes, BostonGlobe, NewZeland, Garvie2016}). The concern is particularly related to positive identification due to the huge consequences of a false match in forensic cases combined with previous failures of face recognition systems in that direction \cite{TheGuardian, Hill2020}.

\begin{figure*}[!h]
	\centering
	\includegraphics[width=0.4\linewidth]{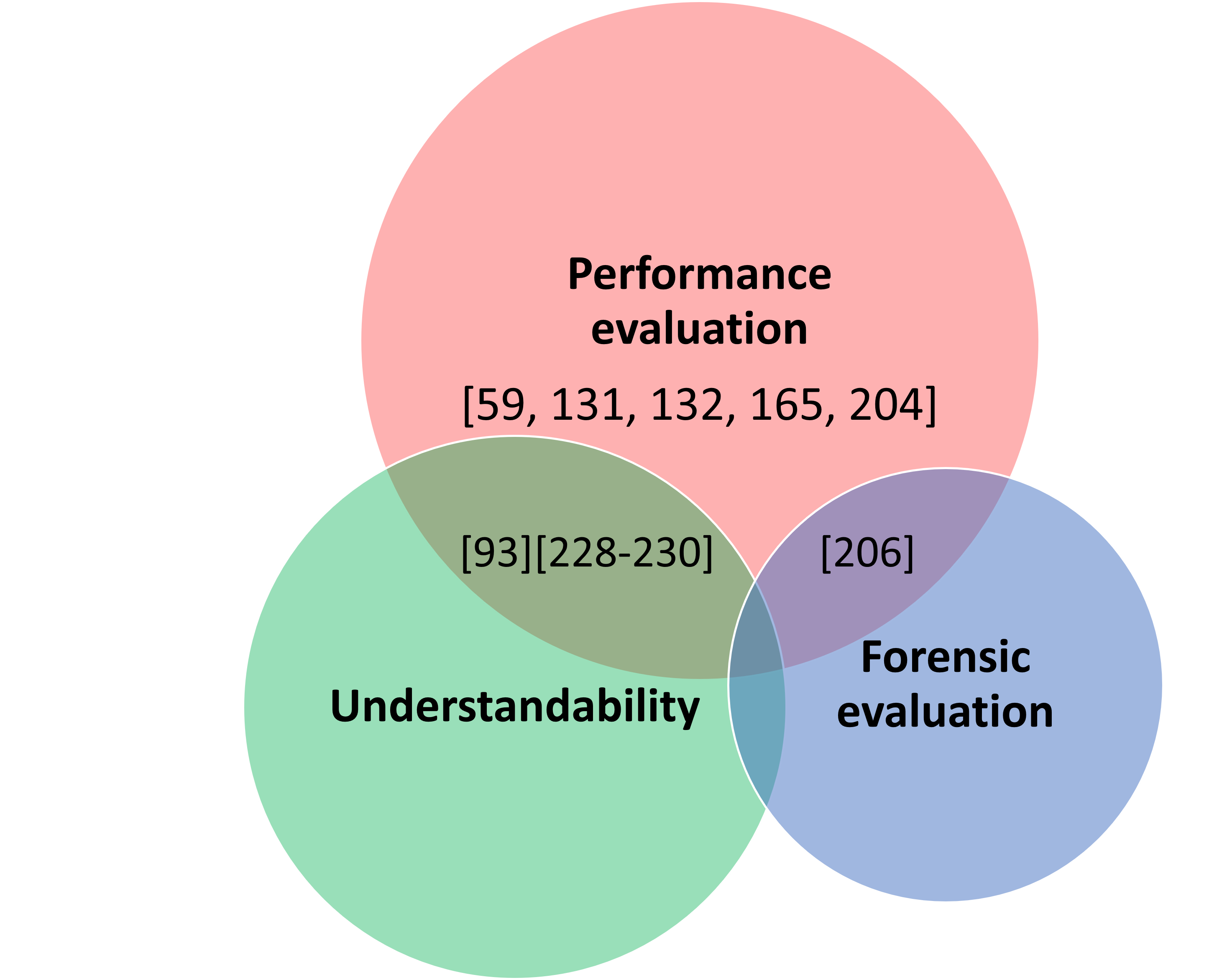}
	\caption{\textit{Taxonomy of forensic recognition methods based on 3DFR with respect to the evaluation levels for forensic purposes.}}
	\label{fig:workseval}
\end{figure*}

Therefore, a robust and transparent methodology must be given for forensic recognition, which effectiveness has to be quantitatively assessable in statistical and probabilistic terms. The goal is to provide guidelines for quantifying biometric evidence value and its strength based on assumptions, operating conditions, and the casework's implicit uncertainty  \cite{Franco2021, Tistarelli2014, NewZeland}. 
Besides, a set of interpretation methods must be defined independently of the baseline biometric system and integrated into the considered algorithm \cite{Neumann2012, Tistarelli2014}. This allows reaching conclusions in court trials in agreement with three constraints (Fig. \ref{fig:workseval}): performance evaluation, understandability, and forensic evaluation \cite{JainSurvey, linkage, forensicRequirements}. Closely related to these constraints, the \textit{quality} of the probe and reference data should also be considered in the admissibility assessment \cite{forensicRequirements, Carabinieri, YanOsadciw2008}.

\subsubsection{Performance Evaluation}\label{subsub:performance}

Performance evaluation concerns the basic trust level of the system and its performance for a specific purpose; therefore, it supports the forensic practitioner's decision when using such a system to perform a given task. For instance, a biometric system could be considered suitable for a specific task whenever it is tested and achieves a performance acknowledged as ``good'' on data representative of the working system's context. \textit{E.g.}, a face recognition system designed to perform well on high-resolution frontal images is not required to achieve the same performance on images acquired by CCTV cameras and random head poses \cite{forensicRequirements}.  
In a statistical evaluation, the definition of ``good performance'' depends on the context, the data and the end users' requirements set in the design process.
The performance parameters are different according to the system itself and the specific task for which it should be employed. For example, the accuracy, namely the percentage of correctly classified samples \cite{AccuracyDefinition}, could be considered in evaluating the performance of classification problems, such as in the case of the face recognition task. 
Distance-based metrics can be instead used for evaluating the error between the predicted values and the real ones in regression problems such as the 3D reconstruction tasks. An example of the latter is the Root Mean Square Error (RMSE), which considers the distance between a reconstructed facial part and the corresponding ground truth in terms of pixels (\textit{e.g.}, \cite{Zeng2017}).
As previously mentioned, understanding the metrics employed requires basic statistics knowledge, which legal decision-makers often do not have. This makes it difficult to justify the use of a particular system by such metrics in a law court \cite{forensicRequirements}. Thus, a certain level of confidence in the underlying technical aspects is necessary to interpret the performance parameters adopted.

Another issue for the trust of biometric recognition systems in forensics is that of biased performance against certain demographic groups, meaning that the performance parameters may depend, on average, on the demographic groups present in the system's data set  \cite{JainSurvey, Wang2019}. For example, biased performance on age, gender, and ethnic groups was recently reported \cite{genderBias, JainSurvey, genderBias2}. In face recognition, bias is a severe problem since facial regions contain rich information strictly correlated to many demographic attributes, which could lead to biased performance \cite{bias}. This issue has often been overlooked when face recognition systems were employed by law enforcement agencies \cite{Grother2011}. Thus, the missed analysis of this aspect, or on the demographic group representative of the casework, could lead to the inadmissibility of the biometric system in judicial trials or, simply, to unreliable support to the human expert decision. 
In other words, the choice of the data sets employed for training the system and evaluating it is one of the factors that must be considered in the performance evaluation \cite{automatedFaces}. 
 Furthermore, fairness, interpretability, and even performance could benefit from the ability of a system to provide information about how biased its decision could be \cite{softMarks, Franco2021} (see also \ref{subsub:understandability}).

\subsubsection{Understandability}\label{subsub:understandability}

Understandability (also known as interpretability \cite{forensicRequirements}) is the ability of a human to understand the functioning of a system, its purpose, its features, as well as its outcome and the (computational) steps that led to such a result. In particular, the understandability evaluation supports the decision of whether the outcome of the system is suitable. This is particularly relevant for legal decision-makers (\textit{e.g.}, judges) who are typically not experts in those topics \cite{AdadiBerrada2018, Arrieta2020, JainSurvey, assessment, forensicRequirements, Gunning2019}. 

A first step for making a system understandable is to design it as ``explainable'' in the decision-making process. This facilitates its traceability, which, in turn, could help prevent or deal with erroneous decisions by revealing the possible points of failure, the most appropriate data and architecture \cite{AdadiBerrada2018, EUEthicsGuidelines, Franco2021}. The main difference between understandability and explainability is that the latter focuses on the system's design \cite{forensicRequirements}, whilst the former focuses on the end-user experience. Therefore, the system's understandability requires an explainable design process.

A factor that can improve the system's understandability is its transparency, meaning the ability of the forensic practitioner to have access to the details related to the functioning of such a system \cite{forensicRequirements}. For example, a fully open-source system is entirely transparent. 
However, even a fully transparent system does not imply its understandability, as in the case of image processing algorithms whose effects cannot be reversed. In other words, they cause a loss of details or an irreversible/random addition which could even impact the reproducibility required by any automated system to be employable by forensic practitioners for reaching conclusions \cite{MARGAGLIOTTI2019138}. Moreover, even details about the algorithms and the implementation of very complex systems like neural networks could be insufficient for their understanding \cite{forensicRequirements}.

Therefore, for both complex and black-box systems, such as those based on Artificial Intelligence, it should be necessary to add sufficient local and/or global interpretations through metrics and mechanisms \cite{AdadiBerrada2018, assessment, forensicRequirements, Franco2021, Guidotti2018, Li2020}. For example, the forensic practitioner must be able to determine whether the system is using the face area instead of the background when computing the related outcome. Moreover, understandability is an aid for legal decision-makers in cases where both the prosecution and the defence of the suspect present contradictory results based on their own black-box systems \cite{JainSurvey}.
Some examples of approaches for enhancing the explainability and, in particular, spatial understandability in the context of face recognition are the extraction of features in different areas of the face \cite{interpretable} and the use of model-agnostic methods (\textit{i.e.}, not tied to a particular type of system \cite{Guidotti2018, AdadiBerrada2018}) that visualizes the salient areas that contribute to the similarity between pairs of faces \cite{modelAgnostic}. Other approaches are the estimation of the uncertainty of features through the analysis of the distributional representation in the feature space of each input facial image, therefore assessing the uncertainty through the variance of such distributions \cite{ShiAndJain}, and the analysis of the effect of features in the resulting outcomes such as facial angle and non-facial elements \cite{Dhar2020, Terhorst2021}. However, black-box systems such as deep neural networks still lack the reasonable interpretability to be effectively employed in forensic processing. In particular, understanding what information is being encoded from the input image into deep face representations would also help address eventual biases of the system (\textit{e.g.}, toward a demographic group) \cite{JainSurvey, frontal2020}. 

\subsubsection{Forensic Evaluation}\label{subsub:forensic}

Forensic evaluation is the assignment of a relative plausibility of information over a set of competing hypotheses (or "propositions") \cite{forensicRequirements}. It supports the forensic practitioner's opinion regarding the level of confidence and the weight (\textit{i.e.}, the strength) of evidence when the system makes a decision according to its outcome \cite{automatedFaces, assessment, forensicRequirements}. The system's performance and understandability are taken into account in forensic evaluation, together with contextual information (\textit{e.g.}, additional cues or supporting evidence from other sources) and general knowledge; thus, additional information which could be either included in the decision process or formalized into the automated system itself \cite{forensicRequirements, marksSurvey, RodriguezLR, misuse}. Therefore, forensic evaluation includes the above elements to drive forensic practitioners toward an appropriate decision (\textit{e.g.}, identification) that could be either conclusive or inconclusive according to the assessed level of confidence \cite{misuse, ColeScheck, Dror, DrorLangenburg, forensicRequirements}.

From a technical perspective, forensic evaluation is quantitatively given by a statistical approach based on the likelihood ratio values (LR) \cite{guidelines, RamosLR, ENFSI, faceslikelihood, Tistarelli2014, Carabinieri}. In particular, it is acknowledged that the LR allows for a transparent, testable, and quantitative assessment of the probability assigned to the evidence of a face match by forensic practitioners, based on personal experience, experiments, and academic research, against the probability of a non-match \cite{forensicRequirements, RodriguezLR}. A semi-quantitative scale could also be employed, in which values are aligned with ranges of likelihood ratios (\textit{e.g.}, weak/medium/strong), or employ the relative strength of forensic observations in light of each proposition \cite{forensicRequirements, ENFSIprimer, relativeLikelihood}. Therefore, thanks to its transparency, testability, and formal correctness, LR allows the clear separation of responsibilities between the forensic examiner and the court. This makes it compliant with the requirements of evidence-based forensic science when quantifying the value of the evidence to the law court \cite{Ramos2017, Saks2005}.
However, it must be remarked that calibrating a biometric score to become an LR requires a substantial amount of case-relevant data, thus data representative of the analyzed scenario regarding quality (see 2.2.4) and demographic group (see 2.2.1).

\subsubsection{Quality Evaluation}\label{subsub:quality}

As previously pointed out, the characteristics of acquired data are also relevant. Firstly, they should meet minimal requirements in terms of \textit{quality} \cite{Carabinieri, YanOsadciw2008}. Although not defined in a rigorous way, this term refers to factors that lead to blurriness, distortion, and artifacts in images. They may be caused by (1) the camera employed, whose sensor, optic, and analog-to-digital converter impact on the image resolution, the dynamic of gray-levels, its ability to focus on the target \cite{automatedFaces, lowquality, clandestine, MalletEvison2013, CrimeScenePhotography, FISWGimage}, (2) environmental conditions such as the illumination and the background of the scene, the same weather conditions (rainy/cloudy) \cite{Introna2010, lowquality, CrimeScenePhotography}, (3) the subject's distance from the camera that adds scaling and out-of-focus problems, his/her camouflage to evade recognition (sunglasses, beard/moustache, hat/cap, makeup, jewelry), the speed at which the subject is moving and the direction, the position of the face with respect to the camera which can lead to non-frontal views and incomplete data \cite{Hu2017, MalletEvison2013, homelandSurvey, Lakshmi2018, positionPerspective, FISWGimage}, (4) the image processing embedded into the camera or next to the raw data acquisition, such as compression and re-sizing  \cite{automatedFaces, lowquality, CrimeScenePhotography, FISWGimage}. Therefore, quality must be evaluated for both probes and reference facial data in order to assess whether the proposed face recognition system is compliant with data of the kind \cite{Introna2010, 2D3Dsurvey, Peng, forensicRequirements}. Secondly, the data amount is crucial from the viewpoint of a new classification system to be trained and fine-tuned \cite{Introna2010, YanOsadciw2008} and the calibration of LR frameworks for the evaluation of those and already existing systems (see 2.2.3), yielding to the creation of large-scale data sets for the evaluation of face recognition algorithms (e.g., \cite{LFW}).

While the acquisition of mugshot images by law enforcement agencies is usually subject to strict control to ensure the truthful representation of appearance, this is often not the case with the acquisition of probe images and videos. Therefore, concerning the available data, it is necessary to assess the quality to determine whether it fulfils the aimed biometric function, including the 3D reconstruction task and the following recognition. The final goal is the system's outcome employment in the forensic investigation and the following judicial conclusion. In the middle, the quality evaluation would allow the assessment of the confidence level in decisions based on such data or to rank and select the ones with the best quality (\textit{e.g.}, single frames from a surveillance video) \cite{schlett2022face, QualityReview, Tome2014, Tome2015}. To the state of our knowledge, a global standard for quality assessment is currently missing \cite{QualityReview, schlett2022face}, probably also due to the human subjectivity factor in the task, and international standards are still under development (\textit{e.g.}, \cite{WDTS24358,JTC1SC37}). However, a score based on the Mean Opinion Scores (MOS) was proposed \cite{CCTVquality} to justify the legal acceptance or the rejection of a potential probe image, video, or part of them. Unfortunately, the MOS method is often impractical since it is considered slow, expensive, and, in general, inconvenient. Although other quality assessment methods have been proposed, most of them are not representative of human perception \cite{perceptionCorrelation, humanPerception}. In our opinion, specific expertise in agreement with the lawcourt process should be included (\textit{e.g.}, \cite{CCTVquality, Tome2014}). 
Furthermore, quality measures about the "partial results" of the system should be integrated as well. For example, in the case of forensic recognition based on 3D face reconstruction from 2D images or videos, the 3D model reconstructed either from reference or probe data could be corrupted due to inaccurate localization of facial landmarks \cite{MediaCollection}, thus requiring the repetition of the localization process or even to discard the sample because it results to be unfeasible.
Therefore, the quality measures could be integrated into forensic recognition, considering them as complementary features  \cite{Fumera, Pisani, QualityReview}.
This means that quality assessment would pass through the previously described requirements (Subsections \ref{subsub:performance}-\ref{subsub:understandability}) in order to be admissible in the analyzed context \cite{schlett2022face} according to the forensic evaluation process (Subsection \ref{subsub:forensic}).

\subsection{3D Face Reconstruction in Forensics} \label{subsec:3DFR}

\begin{figure*}[!h]
	\centering
	\includegraphics[width=0.75\linewidth]{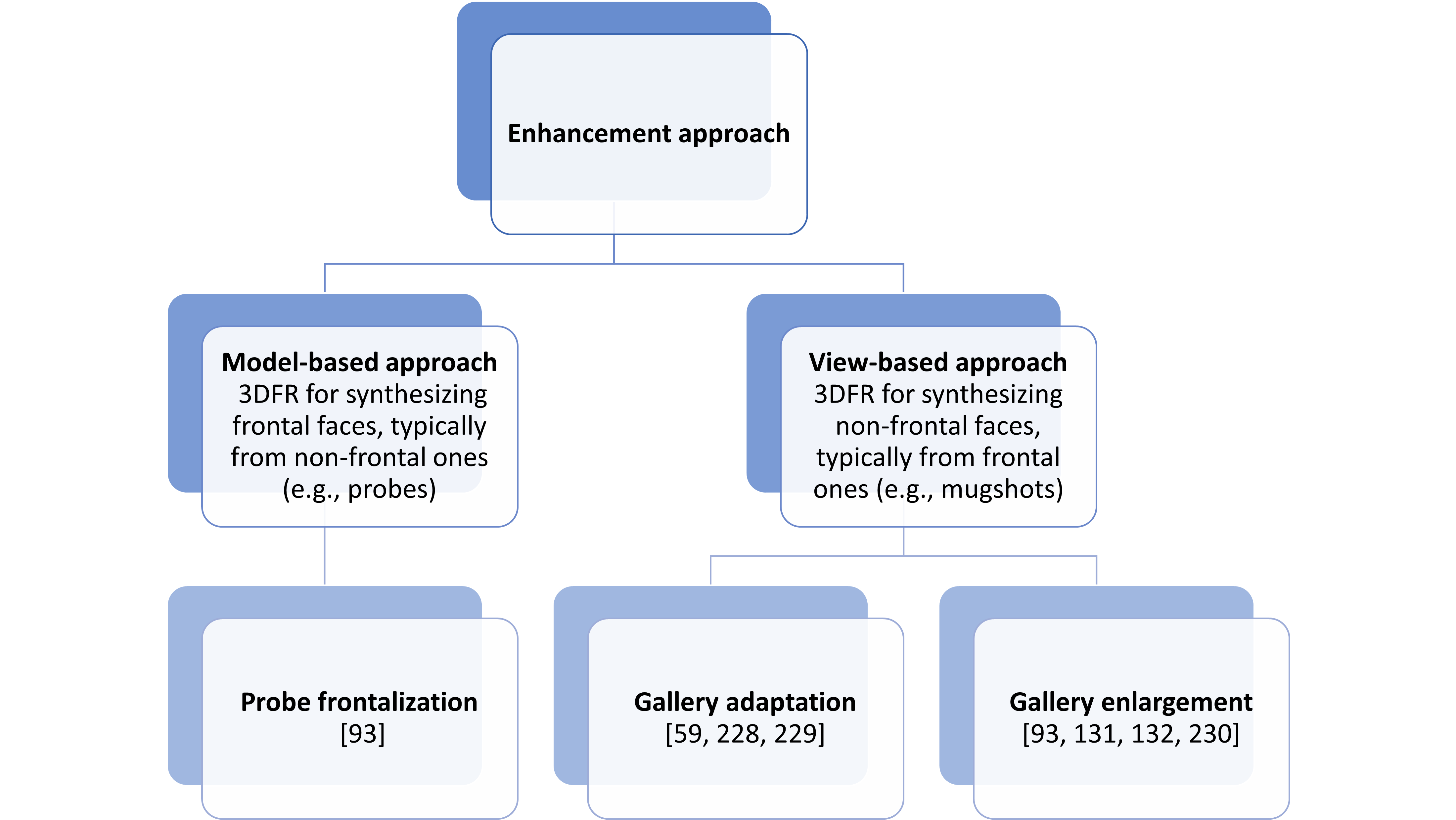}
	\caption{\textit{Taxonomy of performance enhancement methods through 3DFR for forensic recognition.}}
	\label{fig:apptax}
\end{figure*}

During the investigation phase, the subject's identity is unknown, and the possible identities within a suspect reference set need to be rendered and sorted \cite{YanOsadciw2008} in terms of likelihood with respect to the evidence (e.g. a frame captured from a CCTV camera) \cite{Ali2014}. 
In addition to the classic challenges related to facial recognition in uncontrolled environments, such as low resolution, large poses, and occlusions \cite{GuoZhang2019}, forensic recognition faces even more challenges. Examples are the acquisition systems which are set up cheaply and subjects that actively try not to be captured by cameras, which enhance the previously cited issues and introduce novel problems such as heavy compression, distortions, and aberrations \cite{Zeinstra2018}.
Thanks to its greater representational power than 2D facial data, 3DFR can alleviate some of these problems. In fact, 3D data provides a representation of the facial geometry that reduces the adverse impact of non-optimal pose and illumination. 
Depending on the characteristics of the probe image and the reference set narrowed down by police and forensic investigation, whenever the investigator is required to compare these images, and it is necessary or advantageous to use an automatic face recognition system, 3DFR can be employed by following two different approaches, namely \textit{view-based} and \textit{model-based} approaches (Fig. \ref{fig:apptax}), to improve the performance of facial recognition systems and, therefore, enhancing its admissibility in legal trials.

\begin{figure*}[]
	\centering
	\includegraphics[width=0.7\linewidth]{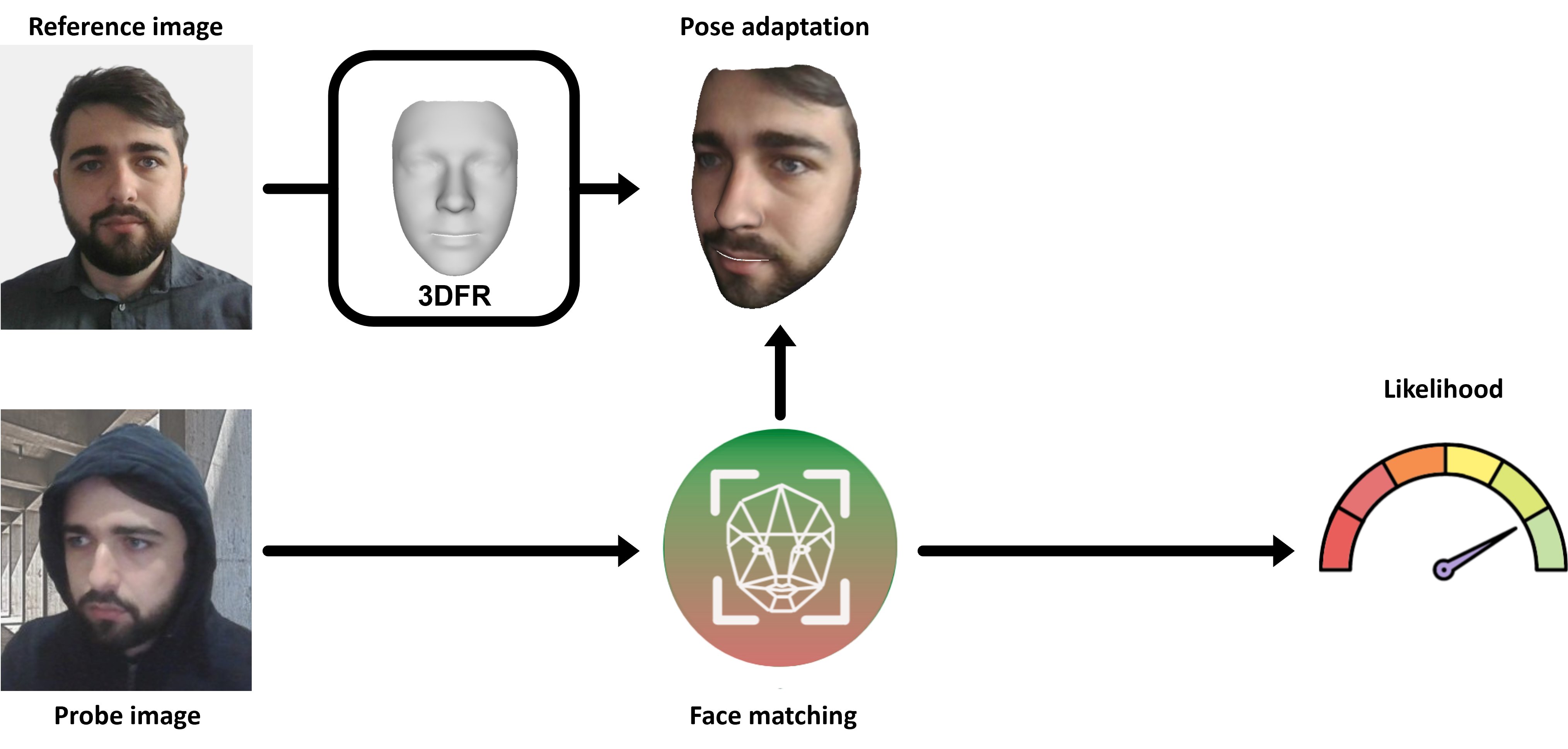}
	\caption{\textit{Example of face recognition through a view-based approach (the 3D model was obtained through \cite{eos}).}}
	\label{fig:viewexample}
\end{figure*}

In a view-based approach, the set of images containing frontal faces is adapted to non-frontal ones, and, thus, it is typically applied on the reference set to adapt the faces within mugshots to the probe image such that it matches the pose of the represented face \cite{view} (Fig. \ref{fig:viewexample}). Although it allows comparing facial images under similar poses, this approach requires a reference set containing images of suspects captured in such a pose or synthesizing such a view through the 3D model of each suspect. In the latter case, each 3D model can be adapted after applying a pose estimation algorithm on the probe image before employing the actual recognition system \cite{Dutta2012, marcialis2014novel, Zeng2016, Zeng2017}. Another proposed strategy is to introduce a gallery enlargement phase instead, which consists of projecting the 3D model in various predefined poses in the 2D domain to enhance the representation capability of each subject and then employing the synthesized images in the recognition task \cite{Zhang2008, HanJain2012, Liang2018, Liang2020}. However, the view-based approach represents a suitable choice whenever multi-view face images of suspects are captured during enrollment for the purpose of highly accurate authentication, such as in the case of the verification task in face recognition \cite{HanJain2012}, although it usually involves higher computational cost in terms of both time and memory with respect to the model-based counterpart.

\begin{figure*}[!h]
	\centering
	\includegraphics[width=0.7\linewidth]{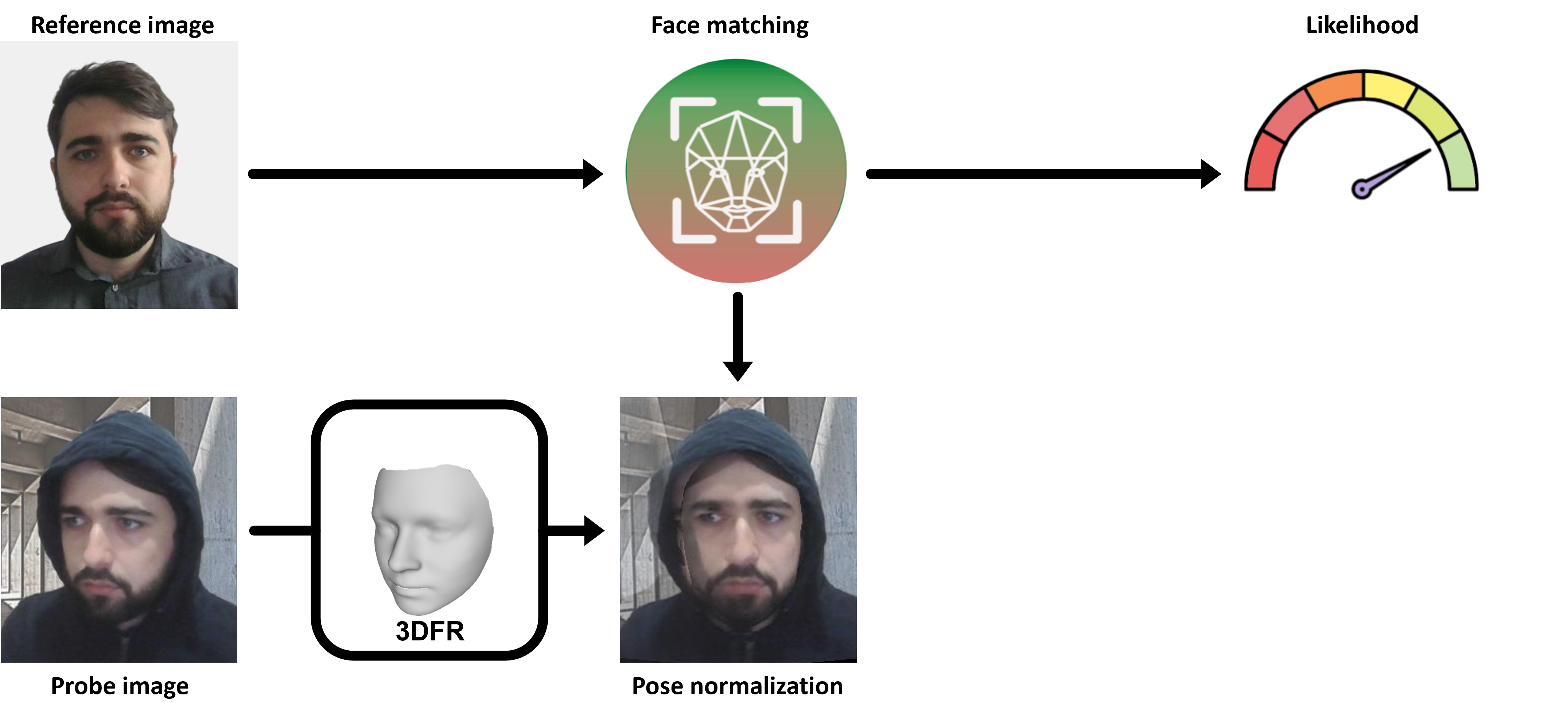}
	\caption{\textit{Example of face recognition through a model-based approach (based on \cite{hassner2015effective}).}}
	\label{fig:modelexample}
\end{figure*}

In a model-based approach, the adaptation phase is performed on non-frontal faces to synthesize a face in frontal view through the reconstructed 3D face \cite{HanJain2012} (Fig. \ref{fig:modelexample}). The normalized (or ``frontalized'') face is then compared to the frontal faces within the gallery set to determine the subject's identity in the probe image \cite{hassner2015effective, Ferrari2016}. This approach is suitable for real-world scenarios in which it is necessary to seek the identity of an unknown person within a probe image or video in a large-scale mugshot data set \cite{HanJain2012}, as in the so-called face identification task in biometric recognition, for maximizing the likelihood of returning the potential candidates.
Despite the generally lower computational cost, this approach is only applicable when it is possible to synthesize good-quality frontal view images with the original texture since it could provide complementary information for recognition with respect to the shape \cite{ICPrecognition, Liu2018}. According to what we discussed in \ref{subsub:quality}, the minimum quality requirements for the probe images must be met, which is not often the case in real forensic scenarios. Furthermore, it could be necessary to handle possible textural artifacts in the resulting frontal image \cite{hassner2015effective, Cao2020, RotateAndRender}.

Hence, the application of a view-based approach would allow changing the scenario from a more traditional 2D-to-2D recognition to a 3D-to-2D recognition, in which the reconstructed 3D face representation is typically used to generate synthetic facial views matched with the probe image \cite{Cadoni2010}. 
This can be achieved by turning the 3D model in such a way that the pose matches the one in the compared image and eventually after applying similar light conditions on the model to ease the comparison (\textit{e.g.}, \cite{wang2003face}). 
Similarly, a model-based approach could be exploited either for aiding the 2D-to-2D face recognition task, through the synthesis of non-frontal faces in the frontal view \cite{HanJain2012}, as it is typically the case of probe images, and the 2D-to-3D recognition scenario, where several synthetic views can provide a set of potential probe images \cite{vanDam2016}, in agreement with the reference ones.
Coherently, these approaches would jointly allow a 3D-to-3D recognition scenario: the 3D representation of the face reconstructed from the reference images is compared with the one reconstructed from a probe video sequence \cite{Cadoni2010}. 
The view-based approach typically involves the reconstruction from mugshots and the model-based approach from probes images, mainly due to the typical qualitative characteristics of data. 
Nonetheless, it is still possible to employ these approaches on both sets of data, according to the specific task (e.g., it could be possible and convenient to apply a view-based approach on a surveillance video to ease the comparison). However, the potential bias towards the average geometry must be taken into account when reconstructing the 3D faces \cite{vanDam2013}, especially when the reconstruction is performed from single images.

\section{3D face reconstruction for mugshot-based recognition}\label{sec:mugshot}

Although many attempts have been performed in the past years to reconstruct faces in the 3D domain, either from a single image or multiple images of the same subject \cite{uncalibrated}, only a few were evaluated for their potential applications in forensics. Among them, we want to focus on exploiting mugshot images captured by law enforcement agencies. The reason is that methods inspired by this approach are closer than others to satisfying the previously seen criteria for their potential admissibility in forensic cases. 

\begin{figure*}[h]
\centering
	\includegraphics[width=0.7\linewidth]{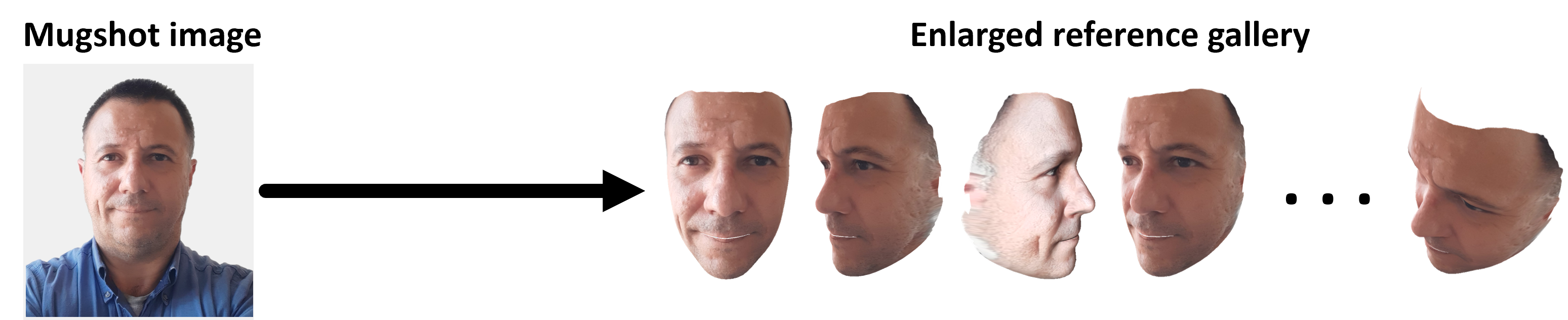}
\caption[justification=centering]{Representation of the gallery enlargement method (the 3D model was obtained through \cite{eos}).}
    \label{fig:enlarge}
\end{figure*}

To our knowledge, the earlier study on 3DFR from mugshot images for forensic recognition was proposed in 2008 by Zhang et al. \cite{Zhang2008}, who employed a view-based gallery enlargement approach to recognize probe face images in arbitrary view with the aid of a 3D face model for each subject reconstructed from mugshot images (Fig. \ref{fig:enlarge}).
To reconstruct the shape of such a model, they proposed a multilevel variation minimization approach that requires a set of landmarks specified on a pair of frontal-side views to be used as constraining points (\textit{i.e.}, eyes, eyebrows, nose profiles, lips, ears, and points interpolated between them \cite{controlpoints}). Finally, they recovered the corresponding facial texture through a photometric method. 
They evaluated their approach on the CMU PIE data set \cite{PIE}, using a holistic face comparator (or matcher) \cite{eigenfaces} and a local one typically employed in biometrics for a textural classification \cite{LBP}, restricting the rotation angles of the probe images to ±70°. This analysis revealed a significant improvement in average recognition accuracy with respect to the original mugshot gallery, especially when the rotation angle of the face in the probe image is larger than 30°.
However, the limit of the rotation angle of faces in probe images and the use of traditional face comparators rather than state-of-the-art ones do not allow for assessing the actual improvement in the effectiveness of 3DFR from mugshot images in terms of forensic recognition \cite{HanJain2012, Liang2018}. Other drawbacks of the proposed method are the possible artifacts caused by the assumed model \cite{Zeng2016} and the poorly explored image texture. Furthermore, they performed the analysis on a small-scale data set containing only 68 subjects. Finally, despite improved performance and the usage of a local face comparator that enhances understandability \cite{interpretable}, expressing the similarity between the single facial parts rather than providing a global similarity and allowing the assessment of the salient areas that led to the outcome of the system, the authors did not utilize any strategy for facilitating the forensic evaluation. Moreover, the analysis of local patterns could also help address the presence of occlusions.
Another aspect that could be considered is the computational time required for the gallery enlargement, which appears to make the method unsuitable for applications having strict time constraints, even considering how old the hardware system on which it has been tested is (Table \ref{tab:mugtable}). We further discussed this factor in Section \ref{sec:conclusion}.

\begin{figure*}[h]
\centering
\includegraphics[width=0.45\linewidth]{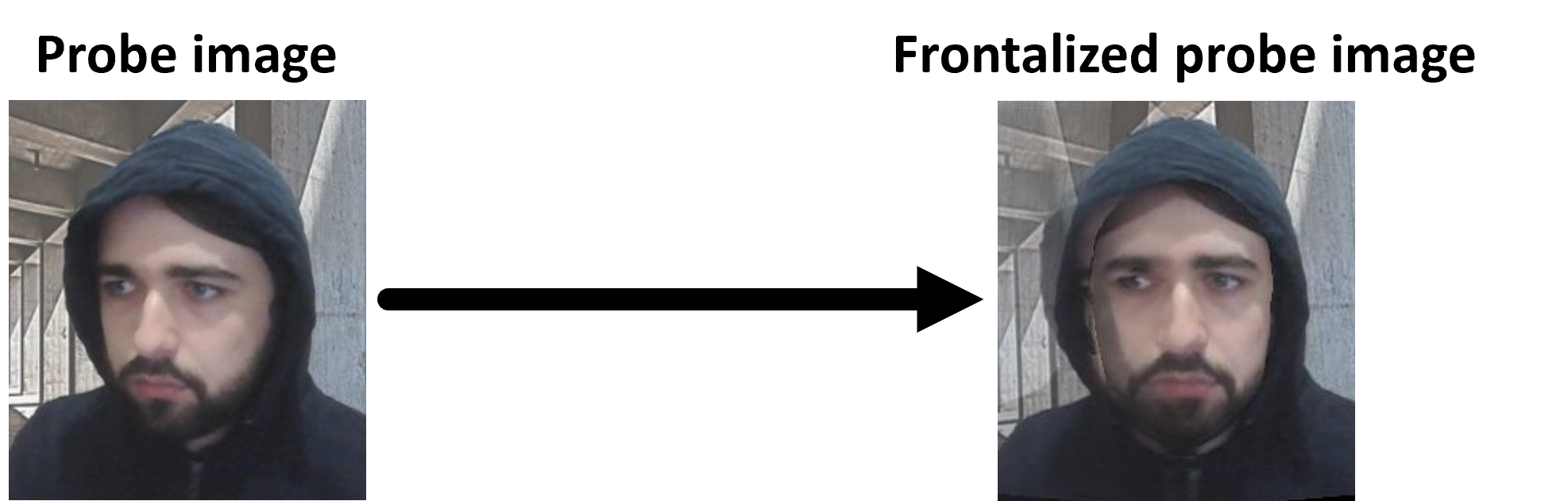}
\caption[justification=centering]{Representation of the probe normalization (or "frontalization") method (based on \cite{hassner2015effective}).}
    \label{fig:frontalize}
\end{figure*}

Four years later, Han and Jain \cite{HanJain2012} proposed to employ the frontalization approach in the considered scenario, as it had already shown its effectiveness in the biometric recognition from non-frontal faces \cite{blanz2005face}. They proposed a 3DFR method from a pair of frontal-profile views based on a 3D Morphable Model (3DMM) \cite{3DMM}, a generative model for realistic face shape and appearance, to aid the reconstruction process. 
They reconstructed the 3D face shape through the correspondence between landmarks within the frontal image and those on the profile one and extracted the texture by mapping the facial image to the 3D shape.
A view-based gallery enlargement approach and model-based probe frontalization approach (Fig. \ref{fig:frontalize}) were employed to enhance the performance through the proposed reconstruction approach. They evaluated them on subsets of PCSO \cite{PCSO} and FERET \cite{ColorFERET} data sets through a local face comparator and a commercial one, revealing an improved recognition accuracy in both cases.
One of the most evident limits of the reconstruction approach in a forensic context is that the involved 3DMM is a global statistical model which is limited in recovering facial details \cite{uncalibrated}, as it could be dominated by the mean 3D face model, which potentially introduces a bias of the outcome towards the underlying model \cite{vanDam2016}. This aspect could be further enforced by the relatively low quality of the employed images. Furthermore, the involved 3DMM could cause evident distortion when the model is largely rotated \cite{Zeng2017, Liang2020}. Other limits of this work are that the authors did not fully explore the texture and did not use state-of-the-art face comparators \cite{Zeng2016, Liang2018}. Therefore, as in the previous case, despite the improvement in performance and the enhanced understandability thanks to local features, the authors did not employ any framework for easing the forensic evaluation of their method. Finally, no information about the computational time was reported.

\begin{figure*}[h]
\centering
\includegraphics[width=0.45\linewidth]{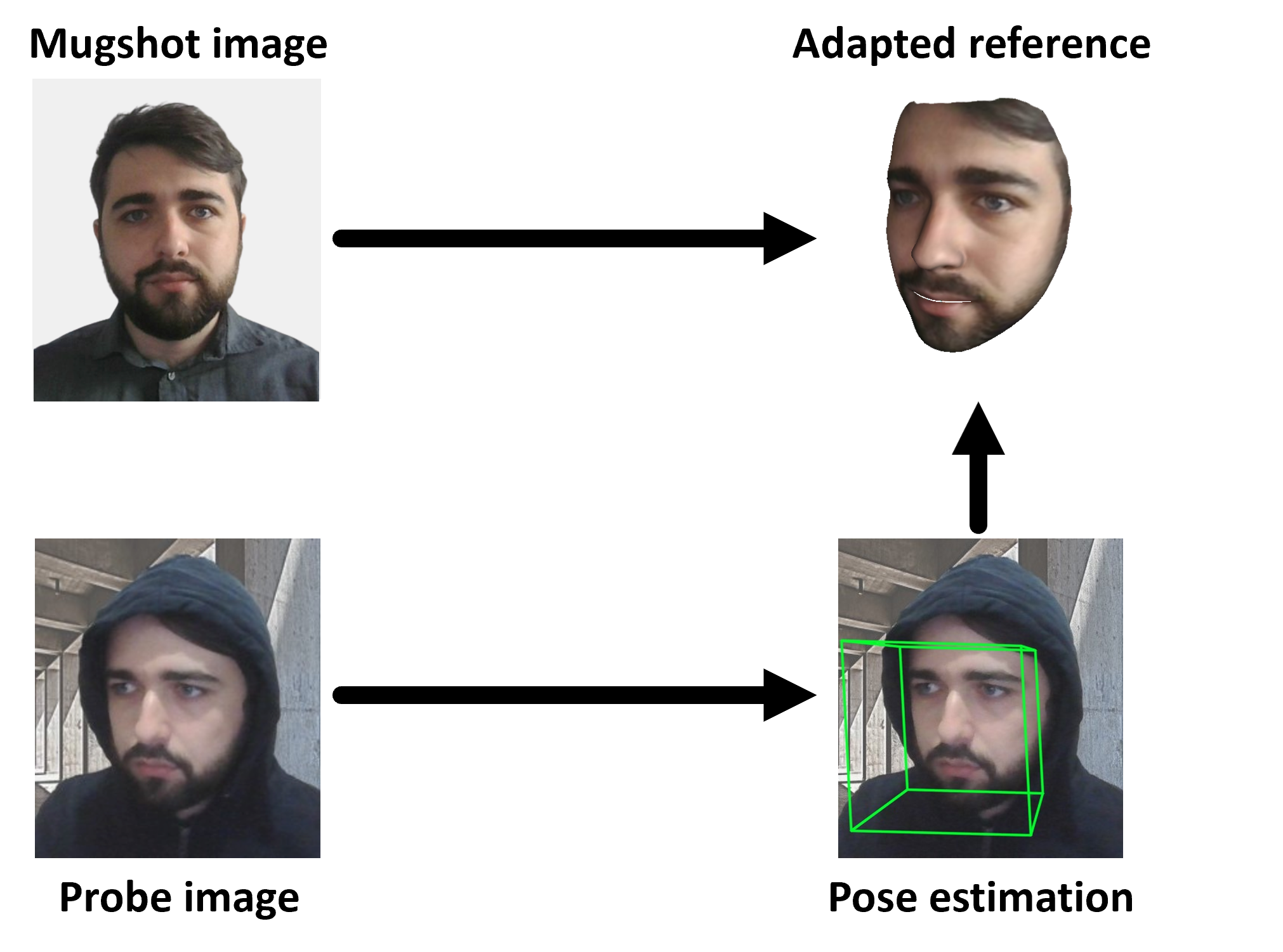}
\caption[justification=centering]{Representation of the gallery adaptation method (the pose was estimated through \cite{guo2020towards,3ddfa_cleardusk}, and the 3D model was obtained through \cite{eos}).}
    \label{fig:adapt}
\end{figure*}

In the same year, Dutta et al. \cite{Dutta2012} proposed a method based on 3DFR for improving face recognition from non-frontal view images through a view-based gallery adaptation approach (Fig. \ref{fig:adapt}).
They applied existing recognition systems to the 16 common subjects in the CMU PIE \cite{PIE} and Multi-PIE \cite{MultiPIE} data sets, containing frontal and surveillance images, respectively.
The adaptation of the reconstructed model to the pose estimated from a probe image could be particularly advantageous whenever poor-quality probe data were acquired, while it is possible to obtain the 3D model from images having a higher quality, such as in the case of mugshot images (Fig. \ref{fig:adapt}). However, this approach requires an accurate estimate of the pose of the face in the probe image.
Furthermore, the small number of subjects involved in the study should be enlarged to simulate a forensic case and evaluate the improvement entity for assessing their applicability in real-case scenarios. Despite the advantages in some application contexts in terms of performance, the authors did not take into account understandability or forensic evaluation. The required computational time was not assessed as well.

Similarly, Zeng et al. \cite{Zeng2016, Zeng2017} reconstructed 3D faces from 2D forensic mugshot images, employing frontal, left profile, and right profile reference images, through multiple reference models to obtain more accurate outcomes for enhancing recognition performance through a view-based gallery adaptation approach. 
To this aim, they used a coarse-to-fine 3D shape reconstruction approach based on the three views through a photometric method and multiple reference 3D face models. The use of multiple reference models is an attempt to limit the homogeneity of reconstructed 3D face shape models and increase the probability of finding the most similar candidate for the single parts of the input face. 
The so-reconstructed 3D face shapes were then used in the recognition task to establish correspondence between the local semantic patches around seven landmarks on the arbitrary view probe image and those on the gallery of mugshot face images, assuming that patches will deform according to the head pose angles. 
The authors \cite{Zeng2016} tested their approach on the CMU PIE \cite{PIE} and Color FERET \cite{ColorFERET} data sets. They showed that deforming semantic patches is effective \cite{LBP} and compared the performance with a commercial face recognition system \cite{VeryLookSDK} and the previously described method proposed by Zhang et al. \cite{Zhang2008}. The authors \cite{Zeng2017} also evaluated the enhancement using a machine learning (ML) classifier on different poses within the Bosphorus \cite{Bos} and Color FERET \cite{ColorFERET} data sets.
As the authors suggested, the improvement in recognition capability from arbitrary position face images is due to the greatest robustness of semantic patches to pose variation and the higher inter-class variation introduced by the subject-specific 3D face model.
A limitation of this work is the out-of-date involved face comparators \cite{Liang2018}. Furthermore, although the method employs multiple reference models, the outcome could still be biased toward them \cite{vanDam2016}.  Finally, despite the fact that the proposed method enhances the performance of an understandable recognition approach, thanks to the employed local recognition approach, the authors did not perform any forensic evaluation. Moreover, despite assessing the test time on a single probe image, the authors did not report the computational time required for the reconstruction of the models in the reference gallery nor for the training of the recognition system (Table \ref{tab:mugtable}).

In 2018, Liang et al. \cite{Liang2018} proposed an approach for arbitrary face recognition based on 3DFR from mugshot images which fully explores image texture.
The proposed shape reconstruction approach is based on cascaded linear regression from 2D facial landmarks estimated in frontal and profile images. After reconstructing the 3D shape, they approached the texture recovery through a coarse-to-fine approach. 
Therefore, they employed the proposed method in a recognition task on a subset of images from each subject of the Multi-PIE data set \cite{MultiPIE} through a view-based gallery enlargement approach on state-of-the-art comparators based on deep learning (DL). Furthermore, they compared the performance before and after the gallery enlargement and by fine-tuning the comparators with the generated multi-view images. The results highlighted improved recognition accuracy in large-pose images, especially with fine-tuned comparators. In particular, this method provides better results than the one proposed by Han and Jain \cite{HanJain2012}, probably because of the major focus on reconstructing texture information \cite{Liang2018}.
Hence, the most significant novelties introduced by this work are the textured full 3D faces reconstructed from the mugshot images and the analysis on DL-based comparators, inherently more robust to pose variations than traditional ones \cite{Liang2018}. Furthermore, they fine-tuned those comparators with the enlarged gallery, revealing even better performance than the previous gallery enlargement approaches.
The authors also assessed the computational time required for the reconstruction of the 3D models, revealing a huge improvement with respect to the previous study reporting it, still considering the different capabilities of the physical system on which it has been tested (Table \ref{tab:mugtable}). Despite the reconstruction method appearing suitable for real-time applications \cite{Liang2018}, the authors did not report the computational time required for training and testing the recognition system.
A limit of the proposed method is that it does not consistently work across all pose directions, revealing worse performance for some poses than in the original gallery (\textit{e.g.}, in frontal pose). Furthermore, the evaluated performance could suffer from demographic bias due to the unbalanced demographic distribution related to the data set employed in the experiments \cite{MultiPIE}. Finally, the authors did not take into account any understandability or forensic evaluation.

In 2020, the same authors published an extension of this work \cite{Liang2020}, in which they also proposed a DL-based shape reconstruction.
In this work, the authors extended the evaluation of the face recognition capability of the proposed method based on linear shape reconstruction by employing a subset of the Color FERET data set \cite{ColorFERET}, obtaining a higher recognition accuracy on average as in the case of the Multi-PIE data set \cite{MultiPIE}.
Furthermore, they tried to solve the drawback of their previous work, related to worse recognition performance for some poses, with respect to usage of the original gallery, through a fusion between the similarity scores obtained by both the original mugshot images and the synthesized ones. The improvements previously observed by combining 2D images and 3D face models in multi-modal approaches \cite{2D3Dsurvey, BowyerSurvey, BowyerSurvey2, ChangSurvey, ChangExperiment} were therefore confirmed. This approach, evaluated on the Multi-PIE data set \cite{MultiPIE}, revealed consistently better performance on all the pose angles. With respect to their previous study, the authors also reported the computational time required for training the recognition system (Table \ref{tab:mugtable}).
Despite the proposed novelties, the authors did not assess if the proposed DL-based shape reconstruction approach is able to enhance recognition capability. Finally, the study did not consider understandability or forensic evaluation.

\begin{figure*}[h]
\centering
\includegraphics[width=0.9\linewidth]{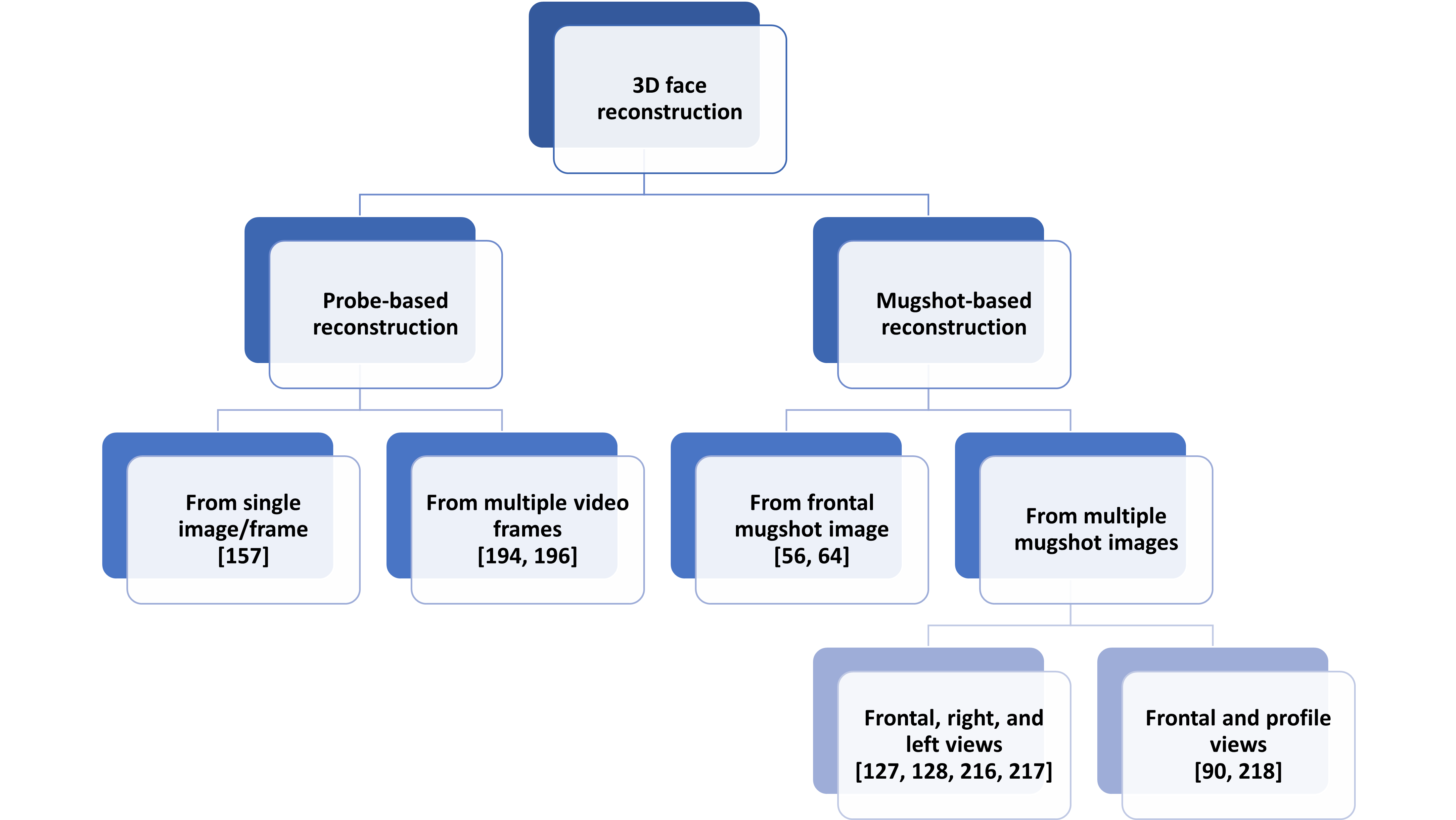}
\caption[justification=centering]{Taxonomy of 3DFR approaches in forensic scenarios.}
    \label{fig:rectax}
\end{figure*}

A quantitative comparison among the previously reviewed methods would require the usage of the same face comparators and their evaluation on the same ground truth data through the same performance metrics, and this is often unfeasible due to many factors, such as the current state-of-the-art data sets when the work has been proposed. Similarly, a comparison in terms of computational time is not suitable both due to the unreported information about time complexity and the differences in terms of physical systems on which the proposed methods have been tested. However, a qualitative comparison is provided in the following table (Table \ref{tab:mugtable}) and then discussed in Section \ref{sec:conclusion}.

\begin{table*}[]
\centering
\caption{Methods based on 3DFR for face recognition from mugshots (FMR is False Match Rate).}
\vspace{-0.35cm}
\scriptsize
\begin{tabular}{|c|c|c|c|cccc|c|c|c|}
\hline
\multirow{4}{*}{\textbf{Method}} &
  \multirow{4}{*}{\textbf{\begin{tabular}[c]{@{}c@{}}Input \\ data \end{tabular}}} &
  \multirow{4}{*}{\textbf{Data sets}} &
  \multirow{4}{*}{\textbf{System}} &
  \multicolumn{4}{c|}{\textbf{Time complexity (s)}} &
  \multirow{4}{*}{\textbf{\begin{tabular}[c]{@{}c@{}c@{}}3DFR \\enhancement \\ approach\end{tabular}}} &
  \multirow{4}{*}{\textbf{\begin{tabular}[c]{@{}c@{}}Performance\\ enhancement\end{tabular}}} &
  \multirow{4}{*}{\textbf{\begin{tabular}[c]{@{}c@{}}Understandability \\ enhancement\end{tabular}}} \\ \cline{5-8}
 &
   &
   &
   &
  \multicolumn{2}{c|}{\textbf{Reconstruction}} &
  \multicolumn{2}{c|}{\textbf{Recognition}} &
   &
   &
   \\ \cline{5-8}
 &
   &
   &
   &
  \multicolumn{1}{c|}{\textbf{Shape}} &
  \multicolumn{1}{c|}{\textbf{Texture}} &
  \multicolumn{1}{c|}{\textbf{\begin{tabular}[c]{@{}c@{}}Training \\ (subjects)\end{tabular}}} &
  \textbf{\begin{tabular}[c]{@{}c@{}}Single \\ test\end{tabular}} &
   &
   &
   \\ \hline
\begin{tabular}[c]{@{}c@{}}Zhang \\ et al.\\ \cite{Zhang2008}\end{tabular} &
  \begin{tabular}[c]{@{}c@{}}Frontal \\and\\ profile\end{tabular} &
  \begin{tabular}[c]{@{}c@{}}CMU PIE \cite{PIE}\end{tabular} &
  \begin{tabular}[c]{@{}c@{}}Intel \\ Pentium IV \\ 2.8-GHz \\ with \\ 1 GB of \\ memory\end{tabular} &
  \multicolumn{1}{c|}{985.8} &
  \multicolumn{1}{c|}{225.6} &
  \multicolumn{1}{c|}{\begin{tabular}[c]{@{}c@{}}1,048.64 \\ (68 \\ subjects)\end{tabular}} &
  0.635 &
  \begin{tabular}[c]{@{}c@{}}Gallery \\ enlargement\end{tabular} &
  \begin{tabular}[c]{@{}c@{}} From 74.27\% to \\ 93.45\% face \\ recognition rate \\ introducing \\ synthesized \\ virtual views\end{tabular} &
  \begin{tabular}[c]{@{}c@{}}Local binary \\ patterns\end{tabular} \\ \hline
\begin{tabular}[c]{@{}c@{}}Han and \\ Jain \cite{HanJain2012}\end{tabular} &
  \begin{tabular}[c]{@{}c@{}}Frontal \\and\\ profile\end{tabular} &
  \begin{tabular}[c]{@{}c@{}c@{}}FERET  \cite{ColorFERET}\\ and PCSO \\ \cite{PCSO}\end{tabular} &
  N.A. &
  \multicolumn{1}{c|}{N.A.} &
  \multicolumn{1}{c|}{N.A.} &
  \multicolumn{1}{c|}{N.A.} &
  N.A. &
  \begin{tabular}[c]{@{}c@{}}Probe \\ frontalization\\ or gallery \\ enlargement\end{tabular} &
  \begin{tabular}[c]{@{}c@{}}From 72.5\% to \\ 82.1\% rank-1 \\ accuracy \\ with probe \\ frontalization,\\ from 0.1\% to \\ 65.5\% verification \\ rate at FMR=0.1\% \\ with gallery \\ enlargement\end{tabular} &
  \begin{tabular}[c]{@{}c@{}}Local binary \\ patterns\end{tabular} \\ \hline
\begin{tabular}[c]{@{}c@{}}Dutta \\ et al.\\ \cite{Dutta2012}\end{tabular} &
  Frontal &
  \begin{tabular}[c]{@{}c@{}}CMU PIE \cite{PIE} \\ and \\ Multi-PIE \cite{MultiPIE}\end{tabular} &
  N.A. &
  \multicolumn{1}{c|}{N.A.} &
  \multicolumn{1}{c|}{N.A.} &
  \multicolumn{1}{c|}{N.A.} &
  N.A. &
  \begin{tabular}[c]{@{}c@{}}Gallery \\ adaptation\end{tabular} & N.A. &
  N.A. \\ \hline
\begin{tabular}[c]{@{}c@{}}Zeng \\ et al.\\ \cite{Zeng2016, Zeng2017}\end{tabular} &
  \begin{tabular}[c]{@{}c@{}}Frontal, \\left\\ profile, \\ and \\ right \\profile\end{tabular} &
  \begin{tabular}[c]{@{}c@{}}Color FERET \\ \cite{ColorFERET}, \\ Bosphorous \\ \cite{Bos}, and \\CMU PIE \cite{PIE}\end{tabular} &
  \begin{tabular}[c]{@{}c@{}}Intel \\ core i5 \\ 2.60-GHz \\ with \\ 4 GB of \\ memory\end{tabular} &
  \multicolumn{1}{c|}{N.A.} &
  \multicolumn{1}{c|}{N.A.} &
  \multicolumn{1}{c|}{\begin{tabular}[c]{@{}c@{}}N.A. (68 \\ subjects)\end{tabular}} &
  9 &
  \begin{tabular}[c]{@{}c@{}}Gallery \\ adaptation\end{tabular} &
  \begin{tabular}[c]{@{}c@{}}Mean accuracy \\of 97.8\%, 2.3\% \\higher than \\Zhang et al. \\ \cite{Zhang2008} \end{tabular} &
  \begin{tabular}[c]{@{}c@{}}Local binary \\ patterns on \\ landmark-based \\ patches\end{tabular} \\ \hline
\begin{tabular}[c]{@{}c@{}}Liang \\ et al.\\ \cite{Liang2018, Liang2020}\end{tabular} &
  \begin{tabular}[c]{@{}c@{}}Frontal, \\left\\ profile, \\ and \\ right \\profile\end{tabular} &
  \begin{tabular}[c]{@{}c@{}}Multi-PIE \cite{MultiPIE}\\ and Color \\ FERET \\ \cite{ColorFERET}\end{tabular} &
  \begin{tabular}[c]{@{}c@{}}Intel \\ core \\ i7-4710 \\ with \\ 16 GB of \\ memory\end{tabular} &
  \multicolumn{1}{c|}{0.04} &
  \multicolumn{1}{c|}{1.1} &
  \multicolumn{1}{c|}{\begin{tabular}[c]{@{}c@{}}133 \\ (1000 \\ samples)\end{tabular}} &
  N.A. &
  \begin{tabular}[c]{@{}c@{}}Gallery \\ enlargement\end{tabular} &
  \begin{tabular}[c]{@{}c@{}}From rank-1 \\ identification \\ rate in the range \\ 70.35-88.22\% to \\ 87.94-94.88\% with\\ DL comparators \\ on Multi-PIE \cite{MultiPIE} \\ (86.30-94.41\% with \\ Han and Jain \cite{HanJain2012})\end{tabular} &
  N.A. \\ \hline
\end{tabular}
\normalsize
\label{tab:mugtable}
\vspace{-0.5cm}
\end{table*}

\section{Other applications of 3D face reconstruction in forensics}\label{sec:other}

In addition to recognition from mugshot images, 3DFR could represent a valuable aid in other forensic contexts to facilitate the recognition of a subject. An example is the search for missing persons. Taking into account such a scenario, Ferková et al. \cite{Ferkova2020} proposed a method that includes demographic information to improve the outcome of the reconstruction from a single frontal image and, at the same time, speed up the related computation. In particular, the method estimates the 3D shape of the missing person's face by taking into account age, gender and the similarity between the landmarks of the reference depth images and those previously annotated in the input image. Then, planar meshes are generated by triangulating between the input image and the depth image. The authors reported that their reconstruction method requires a computational time lower than 3 seconds and strongly depends on the underlying landmarks estimation algorithm.
Despite the good geometrical results, the width of the outcome is usually overstretched, and the generated 3D face model does not include the forehead.
Furthermore, the authors did not quantitatively evaluate the contribution of their method to recognition capability or their potential admissibility in forensic scenarios.

Similarly to some of the previous studies, Rahman et al. \cite{Rahman2016} highlighted how 3D face models could enhance forensic recognition from CCTV camera footage. In particular, they reconstructed the 3D face models from single frames by optimizing an Active Appearance Model (AAM), an algorithm that matches a statistical model of shape and appearance to an image \cite{ABC}. Therefore, they evaluated the improvement in the recognition capability of different ML models with respect to 2D AAMs. However, this study on the possible application of 3DFR to forensic recognition from surveillance videos is limited to a data set of a few subjects, which is not publicly available. Finally, the authors did not assess the recognition performance and did not investigate its admissibility in terms of understandability and forensic evaluation.

With a similar purpose, van Dam et al. \cite{vanDam2012} proposed a method based on a projective reconstruction of facial landmarks. An auto-calibration step is added to obtain the 3D face model from CCTV camera footage. The authors considered the specific case of fraud to an Automatic Transaction Machine (ATM) with an uncalibrated camera under very short distance acquisitions with a distorted perspective \cite{autocalibration}.
They analyzed how the quality of the resulting 3D face model is affected by the number of frames and the number of landmarks, assessing the minimum values for a precise perspective shape reconstruction, which could, however, be affected by the eventual errors on the estimated landmark coordinates introduced by the noise.
However, the authors did not quantitatively assess the method's improvement with respect to its 2D counterpart in face recognition. Neither understandability nor forensic evaluation was addressed.

In 2016, the same authors proposed another method to reconstruct a 3D face from multiple frame images for an application in the forensic context \cite{vanDam2016}. Such a method employs a photometric algorithm to estimate both the texture and the 3D shape of the face. The goal is to avoid generating an outcome biased towards any facial model, thus enhancing the suitability in a forensic comparison process.
The proposed method is a coarse-to-fine shape estimation process: it first provides a coarse 3D shape \cite{vanDam2013} and other pose parameters from landmarks in multiple frames, and then a refined shape is computed by assessing the photometric parameters for every point in the 3D model. The last step also allows estimating the texture information, thus providing the dense 3D model.
The authors evaluated their method in a recognition task on a homemade data set of single-camera video recordings of 48 people containing frames with different facial views. The reconstructed textures with the ground truth images were compared through FaceVACS \cite{FaceVACS} by increasing the considered frames among iterations, revealing enhancement in recognition results in most cases. Furthermore, using the likelihood ratio framework, they highlighted that in more than 60\% of the cases, data initially unsuitable for forensic cases became meaningful in the same context through the proposed method.
As the authors suggested, the outcomes can be used to generate faces under different poses, while they are not suitable for shape-based 3D face recognition.
Despite the enhanced suitability in forensic scenarios, one of the most significant drawbacks is that the model-free reconstruction approach is computationally more burdensome than a model-based one and requires multiple images. Furthermore, the authors did not quantitatively evaluate their method on publicly available data sets. Although the authors did not assess understandability, they introduced a forensic evaluation of their method based on 3DFR; thus, in our opinion, this is the most significant work on 3DFR applied to forensics.

Unlike all previous approaches, Loohuis \cite{Loohuis2021} proposed to employ 3DFR for facing the lack of facial images, which could be used in training ML and DL models for face recognition tasks, for example, in a surveillance scenario. The author combined a method for generating face images with rendering techniques to simulate such adverse conditions and assessed the impact of the resulting synthetic images on existing face recognition systems. In particular, the method proposed by Deng et al. \cite{Deng2019} for reconstructing the 3D model of the face, based on a DL model \cite{he2016deep} and a 3DMM \cite{Basel}, has been applied to the single images of a subset of the ForenFace data set \cite{ForenFace} to generate images simulating different levels of image degradation.
Unfortunately, the proposed method does not perform well on very low-quality images. However, a reasonable level of degradation in many forensic scenarios can still be mimicked because the generated images show a high degree of similarity with the reference ones. Moreover, a similar approach employing 3DFR for generating degraded synthetic views has already been demonstrated to enhance the recognition performance of automatic face recognition systems from low-quality videos, such as those acquired by surveillance cameras, with holistic, local, and DL approaches \cite{Hu2017}.
Furthermore, despite the human subjectivity in perceiving the quality of an image, such an approach could even be employed in the development of quality assessment algorithms for facial images since it would allow comparing the degraded image against a known reference version thereof, thus aiding the selection of potentially suitable samples either for the reconstruction or the recognition tasks \cite{schlett2022face}.

\section{Data sets for face recognition based on 3D face reconstruction} \label{sec:databases}

Public data sets provide a way to test and compare the performance of face recognition systems through a common evaluation framework. Therefore, in this Section, we focus on the characteristics of the available data sets from the perspective of an application of forensic facial recognition based on 3D face reconstruction. Some of them have already been introduced in Sections \ref{sec:mugshot} and \ref{sec:other}.
Furthermore, we provide some proposals about not yet employed data sets, in our opinion, suitable for the forensic task. 

We subdivided the available data sets into two categories. The first one (Subsection \ref{subsec:imagedata}) is related to sets of 2D images containing mugshot-like facial images and, eventually, in-the-wild images (\textit{i.e.}, images acquired in an uncontrolled environment). These include data to test reconstruction algorithms and recognition methods in realistic forensic scenarios. 
The second category (Subsection \ref{subsec:videodata}) includes sets of 3D facial scans and videos, which could be employed to evaluate the accuracy of the 3D reconstruction algorithm and eventual 3D-to-2D or 3D-to-3D face recognition systems, thus extending the application scenarios to realistic surveillance videos.
\vspace{-0.1cm}
\subsection{Image data sets} \label{subsec:imagedata}
\vspace{-0.05cm}
Five different data sets containing 2D images were employed in the previous studies. These data sets contain either RGB images (\textit{i.e.}, color images) or grayscale images acquired in controlled or semi-controlled scenarios. Consequently, most of them could be considered mugshot-like data sets (Table \ref{tab:mugdb}) and employed in studies related to mugshot-based face recognition (Section \ref{sec:mugshot}). However, some of them also contain facial images in different poses and expressions, suitable for evaluating the robustness of the proposed algorithms to such factors. We provided their description and suggested some of their possible uses in studies related to forensic face recognition based on 3D face reconstruction. Besides these, we indicated data sets unemployed in the previous studies but of great potential in our view, allowing us to address some shortcomings of the other data sets or even being explicitly designed for realistic forensic scenarios.

\begin{table*}[htbp]
\vspace{-0.2cm}
\caption{Mugshot-like data sets}
\vspace{-0.35cm}
\scriptsize
\begin{center}
\begin{tabular}{|c|c|c|c|c|c|}
\hline
\textbf{Data set} & \textbf{Image types} & \textbf{Subjects} & \textbf{Forensic features} & \textbf{Acquisition context} & \textbf{Used by}\\
\hline
Color FERET \cite{ColorFERET} & RGB & 994 & None & Semi-controlled & \cite{Zeng2016}\cite{Zeng2017}\cite{Liang2020}  \\
FERET \cite{ColorFERET} & Grayscale & 1199 & None & Semi-controlled & \cite{HanJain2012}  \\
CMU PIE \cite{PIE} & RGB & 68 & None & Controlled & \cite{Zhang2008}\cite{Dutta2012}\cite{Zeng2016}  \\
Multi-PIE \cite{MultiPIE} & RGB & 337 & Landmarks & Controlled & \cite{Dutta2012}\cite{Liang2020}  \\
PCSO \cite{PCSO} & RGB & 28557 & None & Controlled & \cite{HanJain2012}  \\
NIST MID \cite{NISTmug} & Grayscale & 1573 & None & Controlled & \\
Morph (Academic) \cite{MorphAcademic} & RGB & 13618 & Eye coodinates & Controlled & \\
SCface \cite{SCFace} & RGB \& IR & 130 & Landmarks & Controlled \& uncontrolled & \\
ATVS Forensic DB \cite{ATVS} & RGB & 50 & Landmarks & Controlled & \\
LFW \cite{LFW} & RGB & 5749 & None & Uncontrolled & \\
\hline
\end{tabular}
\label{tab:mugdb}
\end{center}
\normalsize
\vspace{-0.4cm}
\end{table*}

The \textbf{Color FERET} data set \cite{ColorFERET} contains multi-pose, multi-expression, and multi-session facial images captured in a semi-controlled environment during 15 sessions across nearly three years, intended to aid in the development of the forensic field. It contains RGB images of size 512×768 pixels. The face of each individual was captured in up to 13 different poses and sometimes on different dates, with an average of about 11 samples per subject. These images represent the frontal pose with different facial expressions, the right and left profiles at different angles with respect to the frontal one, and extra irregular positions. Furthermore, some images were captured while individuals were wearing eyeglasses or pulling their hair back, adding further intra-subject variability to the samples. 
This data set has been analyzed by some of the previously reviewed studies \cite{Zeng2016, Zeng2017, Liang2020} for its application to biometric recognition based on 3D reconstruction from multi-view facial images, considering them as mugshots, while using other samples of the same subjects for evaluating the recognition performance and the robustness of the system to pose and facial expression.
Despite the absence of entirely uncontrolled acquisition, the variations in scale, pose, expression, and illumination, together with the relatively low quality of the images, make the data set potentially suitable for studies related to 3D face reconstruction for surveillance-related tasks, such as the mugshot-based recognition of a suspect captured by a CCTV camera. Furthermore, its multi-session characteristic makes the data set even suitable for studies related to aging, in order to make the system more robust to changes in the appearance of a person's face over time \cite{2DDatabasesSurvey}.

The data set referenced as \textbf{FERET} is the grayscale version of the Color FERET data set. It has been used for evaluating the recognition performance of 3D face reconstruction based on a pair of frontal-profile facial images \cite{HanJain2012}. Since the images are grayscale, the potential applications of this data set appear limited with respect to its colorized version, significantly reducing information about the appearance of the subjects, while the main advantage of this data set is the lower memory required, since a single color channel is used. 

The \textbf{CMU-PIE} data set \cite{PIE} is made up of multi-pose, multi-expression and multi-illumination face images. It contains 41368 RGB images of size 640x486 pixels, collected through a common setup composed of 13 fixed cameras and 21 flashes. Therefore, the faces of individuals were acquired in up to 13 different poses, under 43 different illumination conditions, and with four different expressions. Furthermore, a background image from each of the 13 cameras was acquired at each recording session to ease face localization.
Subsets of CMU-PIE were involved in some of the previously described studies, aiming to evaluate how the recognition performance could benefit from the 3D face reconstruction obtained from a single frontal image \cite{Dutta2012}, a pair of frontal-profile images \cite{Zhang2008}, or the frontal image and both left and right profile images \cite{Zeng2016}, evaluating the robustness of the system to large poses. Furthermore, due to its nature, CMU-PIE can also be used for evaluating the robustness of such systems to illumination conditions and facial expressions.
The controlled environment could limit the usage of a system based on this data set, making it unsuitable for in-the-wild applications. However, its multi-camera setting makes it possible to obtain more images of the same subjects with the same environmental condition as in a registration-like scenario, eventually making the 3D reconstruction from multiple images more straightforward and aiding detailed geometric and photometric modeling of the faces \cite{PIE, Georghiades}. Other limits of this data set are the relatively low number of subjects (Table \ref{tab:mugdb}), which represents a shortcoming for evaluating the inter-subject discriminability, and the limited intra-subject variability due to the single-session scenario and the small range of expressions \cite{MultiPIE}.

The \textbf{CMU Multi-PIE} data set was collected to address such issues \cite{MultiPIE}. In particular, it provides 755370 facial RGB images of size 3072×2048 pixels collected by 15 cameras using 18 different flashes in a controlled environment, with similar settings used for collecting the CMU PIE data set \cite{PIE}. Hence, it represents a multi-pose, multi-expression, multi-illumination and multi-session face data set, which introduces more variability thanks to up to 6 facial expressions and four sessions, increasing the quality of the images as well. 
This data set has been used to evaluate a recognition system's performance based on 3D face reconstruction from frontal and profile images \cite{Liang2020}. Furthermore, it has been used to evaluate the biometric performance on non-frontal images of 16 subjects common with the CMU-PIE \cite{PIE}, from which the 3D face was reconstructed to synthesise a non-frontal view of the subject, which can be compared with the tested image \cite{Dutta2012}.
Hence, in addition to appearing, on the whole, suitable for the tasks already described while discussing the CMU-PIE \cite{PIE}, CMU Multi PIE offers the possibility to evaluate the aging robustness \cite{2DDatabasesSurvey}, especially jointly with its predecessor, which was acquired about four years earlier, even if in a limited way due to the limited number of common subjects. These would also allow the evaluation of performance by employing different acquisition parameters. 
Another interesting feature of this data set is the presence of 68 annotated facial landmark points for images in the range 0° to 45° both left and right and 39 points for profile images, which could be exploited in both reconstruction and recognition algorithms.
The most evident disadvantage of this data set is still the collection in a strictly controlled environment, which makes it unsuitable for in-the-wild applications. Finally, its demographic distributions could lead to a bias since the subjects were predominantly men (69.7\%) and European-Americans (60\%) \cite{MultiPIE}.

The \textbf{PCSO} data set contains mugshot images collected as part of the booking process. One or more RGB images per subject are completed with metadata such as age, sex, and ethnicity. Despite some variations in lighting conditions and head positions, photographic parameters are relatively consistent, and the quality of the images is quite good, with good contrast between the background and the individual, who is photographed with a frontal face and neutral expression \cite{PCSOdesc}. 
The most significant advantage of this data set with respect to the previously described ones is its large number of subjects (Table \ref{tab:mugdb}), making it suitable for longitudinal research.
However, there is no intuitive way to relate multiple arrest records from the same individual \cite{PCSOdesc}, making it challenging to perform multi-session analyses. Moreover, it appears to be unsuitable for studies on robustness in the wild due to its semi-controlled nature. Finally, the data set does not appear to be currently available to other researchers except those already enabled in the past \cite{PCSOna}.

Due to this availability issue, a possible alternative to the PCSO  is the \textbf{NIST-MID} \cite{NISTmug}, containing frontal and profile facial views. Although some subjects were not acquired in the profile view, other subjects were acquired even more than once in both frontal and profile views. However, these mugshot images are in 8-bit grayscale and, therefore, do not allow fully exploiting the information provided by the face's texture. Furthermore, the ratio between male and female subjects (\textit{i.e.}, about 19 males for each female) could lead to a demographic bias. 

Another alternative is the academic version of the \textbf{MORPH} data set \cite{MorphAcademic}, which contains scanned frontal and profile mugshot images related to 13618 subjects \cite{MorphRelease}. 
The images were acquired in different periods of time, up to 1681 days, with an average longitudinal time between photos of 164 days and an average of 4 acquisitions per subject \cite{wrinklesSurvey}, thus allowing the evaluation of the robustness of a recognition system to time progression. MORPH also allows analyzing the system's robustness to the age variation across different subjects since ages range from 16 to 77. 
Moreover, it also contains annotations related to the location of the eyes \cite{ForenFace}, which could be required by some recognition algorithms or employed for evaluating their automatic detection. However, the algorithms tested on this data set could also suffer from a demographic bias due to unbalance between male and female individuals and in terms of ethnicity.

A data set simulating realistic forensic scenarios is the \textbf{SCface} \cite{SCFace}, providing both mugshot and surveillance images acquired through a high-quality photo camera in controlled conditions and five different commercial cameras at the same height, respectively (\textit{e.g.}, Fig. \ref{fig:mugshotvsprobe}). NIR (near-infrared) mugshot images are included as well. The probe images were acquired indoors using the outdoor light coming through a window on one side as the only illumination source. The observed head poses are the ones typically found in footage acquired by a regular commercial surveillance system, with the camera placed slightly above the subject's head \cite{metadataSurvey}.
In total, 21 images of each subject were taken at three different distances from each surveillance camera, between 1 and 4.2 meters. The RGB mugshot images were also cropped by following the ANSI 385-2004 standard recommendations \cite{ANSI2004}. Furthermore, SCFace provides 21 manually annotated facial landmarks \cite{SCFaceAnnotations}, which could be employed in both reconstruction and recognition algorithms and metadata on demographic information and the presence of glasses and moustache \cite{metadataSurvey}. Therefore, this data set could be suitable for studies on photoanthropometry.
To summarize, SCFace could be employed to analyze the effect of different quality and resolution cameras on face recognition performance and the robustness to different illumination conditions, distances, and head poses. SCFace also allows studies on recognition from NIR images, which are inherently more robust to illumination changes than images acquired in the visible spectrum \cite{databasesSurvey}. However, this aspect could find limited application in real-world scenarios due to the specific hardware system required to acquire NIR images \cite{NIR}. From a 3D reconstruction perspective, the information provided by the nine different poses makes this data set also suitable for evaluating the performance of the related algorithms in a realistic scenario.
Although its characteristics make it suitable for low-resolution face recognition in forensic research, traces in the SCface data set only consist of frontal surveillance camera images \cite{ForenFace}. Furthermore, the difference in the distribution between male and female individuals (\textit{i.e.}, 114 and 16, respectively) and the absence of non-Caucasian people could lead to a demographic bias.

Another set of images containing forensic annotations (\textit{i.e.}, 21 landmarks on frontal faces) is the \textbf{ATVS-Forensic} \cite{ATVS}. Despite the relatively low number of subjects (\textit{i.e.}, 32 men and 18 women) and the limitation concerning the potential application scenarios since the data set only consists of high-quality mugshot images, this data set would allow evaluating the robustness of the recognition system to the distance thanks to the acquisition at three different distances between 1 and 3 meters from the camera. Furthermore, it provides a lateral view of the full body and the face. All the images were acquired in each of the two sessions held for each subject, therefore simulating forensic scenarios in which the mugshot images of a suspect have been acquired on a different day with respect to the probe image.

One mention that must be made is that of the \textbf{LFW} data set \cite{LFW}. Although not explicitly designed for forensic applications, it has been employed in many face recognition algorithms that can cope with uncontrolled settings. It contains images acquired in unconstrained scenarios, including variations of pose, expression, hairstyles, camera parameters, background, lightning, and other demographic aspects. Due to the variability in terms of the number of images for each subject, up to 530, LFW is suitable for both identification and verification scenarios.

The considerable differences among the reviewed data sets make them suitable for different purposes. Therefore, future studies should consider these differences in order to assess whether a specific data set is suitable for the performance evaluation of the proposed system, starting from the representativeness of the images contained with respect to the aimed scenario.
\vspace{-0.1cm}
\subsection{3D scan and video data sets}
\label{subsec:videodata}
Despite the smaller amount of available data sets, videos and 3D face scans could be effectively employed in order to evaluate the proposed 3D face reconstruction algorithms. Their characteristics are summarized in Table \ref{tab:scandb}. In particular, the acquisition context of the analyzed dataset could make them suitable for different scenarios that are characteristics of the forensic fields, either in terms of reference images or probe data. Moreover, most of them contain annotations which are traditionally employed in forensic cases (\textit{e.g.}, landmarks).
These features motivate their potential in the simulation of the face comparison from surveillance footage. 

\begin{table*}[htbp]
\vspace{-0.2 cm}
\caption{3D scan and video data sets}
\vspace{-0.35cm}
\scriptsize
\begin{center}
\begin{tabular}{|c|c|c|c|c|c|}
\hline
\textbf{Data set} & \textbf{Data types} & \textbf{Subjects} & \textbf{Forensic features} & \textbf{Acquisition context} & \textbf{Used by}\\
\hline
Bosphours \cite{Bos} & 3D scans \& images & 105 & Landmarks & Controlled & \cite{Zeng2017, Liang2020}  \\
ForenFace \cite{ForenFace} & 3D scans, videos \& images & 97 & Annotated facial parts & Controlled \& uncontrolled & \cite{Loohuis2021}  \\
Quis-Campi \cite{QuisCampi} & 3D scans, videos, images \& gait & 320 & Eye coordinates & Controlled \& uncontrolled & \\
Wits Face Database \cite{Wits} & Videos \& images & 622 & None & Controlled \& uncontrolled & \\
IJB-C \cite{8411217} & Videos \& images & 3531 & None & Uncontrolled & \\
FIDENTIS (Licensed) \cite{FIDENTIS} & 3D scans & 200 & Landmarks & Controlled & \\
NoW benchmark \cite{RingNet:CVPR:2019} & 3D scans \& images & 100 & None & Controlled \& uncontrolled & \\
Florence 2D/3D \cite{bagdanov2011florence} & 3D scans \& videos & 53 & None & Controlled \& uncontrolled & \\
\hline

\end{tabular}
\label{tab:scandb}
\end{center}
\normalsize
\vspace{-0.35 cm}
\end{table*}

The \textbf{Bosphorus} data set \cite{Bos} contains both multi-pose and multi-expression 3D data representing the shape of the face and the correspondent RGB texture images of size 1600x1200 pixels. It comprises 4666 face scans related to 60 men and 45 women, mainly Caucasian and aged between 25 and 35. The scans were acquired in a single view using a structured-light-based 3D system while the subjects were sitting at a distance of about 1.5 meters. Several face scans are available per subject, in up to 13 head poses with different yaw and pitch angles, and up to 4 deliberate occlusions of eyes or mouth through beard, moustache, hair, hand or eyeglasses, and 34 different facial expressions for each. Furthermore, it provides up to 24 facial landmarks manually annotated on 2D and 3D images, making it suitable for studies based on photoanthropometry and the estimation of landmarks.
Bosphorus was involved in the evaluation of the performance of a recognition system based on the 3D face reconstruction from multi-view facial images \cite{Zeng2017} and the assessment of the accuracy of the reconstruction from frontal and profile images \cite{Zeng2017, Liang2020}.
This data set is suitable for studies on robustness to occlusions and adverse conditions such as different poses and expressions, thanks to its big intra-subjects variability. One of its main disadvantages is the low ethnic diversity \cite{BosSurvey}. Moreover, the acquisitions under uniform illumination do not allow investigation of the effects of the light variations on reconstruction and recognition. Finally, it contains corrupted data due to subject movements during acquisitions and self-occlusion.

The \textbf{ForenFace} data set \cite{ForenFace} contains 3D scans, videos, and high-quality mugshot images and has been specifically designed to represent realistic forensic scenarios. In particular, ForenFace contains images of five views per subject, photos of an identity document (\textit{i.e.}, employee cards taken months or even years before), and the related frontal and semi-profile 3D scan as reference material. The CCTV videos and stills from visible and partially occluded subjects were instead acquired indoors through six different models of surveillance cameras in various locations, positions, and distances from the subject. ForenFace also includes a large set of anthropomorphic features that forensic facial practitioners employ during forensic work, such as those proposed by FISWG \cite{FISWGmorph}; this makes it suitable for studies on morphological comparison and valuable due to the lack of data sets of facial features allowing quantitative, statistical evaluation of face comparison evidence \cite{Moreton2021}.
This data set is very flexible in its usage and suitable for studies related to various application scenarios. For example, it is suitable for evaluating errors with different models of surveillance cameras. Another potential use is in evaluating the robustness of age differences through passport-style images. The acquired videos allow for assessing the robustness to partial occlusions of the face, thanks to eyeglasses, beard, and baseball caps, and evaluating the reconstruction from probe videos and frontal/profile images. It could also be employed to evaluate methods for extracting facial features and comparing them with the annotated ones. Finally, ForenFace allows the recognition task across different types of facial data (\textit{e.g.}, probe video vs mugshot image or 3D scan).
Despite being particularly useful for forensic research, the size of ForenFace is relatively small from a biometric perspective \cite{ForenFace}. Furthermore, the predominance of the Caucasian ethnicity could lead to demographic bias.

The \textbf{Quis-Campi} \cite{QuisCampi} data set is made up of videos and images taken from modern surveillance systems that typically have a higher resolution than traditional ones. Compared to the previous data set, Quis-Campi contains data related to more subjects (Table \ref{tab:scandb}) captured in the outdoor environment in unconstrained conditions through a camera about 50 meters from the subject. It also contains 3D scans of the face and reference images acquired indoors. Furthermore, it provides gait recordings as full-body video sequences, which could be employed in a multimodal recognition system. Annotations about the locations of the eyes in each frame were also added, which can be useful for evaluating the performance of eye detection or head-pose estimation algorithms.
In summary, Quis-Campi can be adopted to assess the robustness to the key adverse factors of forensic face recognition in the wild, namely expression, occlusion, illumination, pose, motion-blur, and out-of-focus, in a realistic outdoor scenario through an automated image acquisition of a non-cooperative subject on-the-move and at-a-distance \cite{QuisCampi}.
On the other hand, it lacks a good set of reference images \cite{ForenFace} and could lead to demographic bias due to the predominance of the Caucasian ethnicity.

In order to perform a CCTV-based recognition, including both the identification and the verification scenarios, it is possible to employ the \textbf{Wits-Face} data set \cite{Wits}. It includes African male individuals aged between 18 and 35, each acquired in ten photos, in five different frontal and profile views with a neutral expression and facing straight ahead, both under natural outdoor lighting and artificial indoor fluorescent lighting conditions. CCTV video recordings were acquired from 334 subjects in indoor or outdoor environments, allowing the evaluation of the difference in a face recognition system's performance between these.
Furthermore, some of the recordings are related to subjects wearing caps or sunglasses, thus allowing the evaluation of the robustness of such partial occlusions.
One critical issue of Wits-Face is related to demographic bias since only images and videos related to male subjects were acquired.

In the context of face recognition from videos, it is worth mentioning the \textbf{IARPA Janus Benchmark (IJB)} data sets, which contain facial images and videos varying in pose, illumination, expression, resolution, and occlusion, mainly acquired in an uncontrolled scenario. The most recent of these data sets is the \textbf{IJB-C} \cite{8411217}. In particular, it provides 21294 facial images and 11779 face videos of 3531 subjects. Furthermore, all media has manually annotated facial bounding boxes, and the data set includes 10040 non-facial images, allowing studies related to face detection as well. Finally, attribute metadata related to age, gender, occlusion, capture environment, skin tone, facial hair, and face yaw is provided as well, allowing further examinations like occlusion detection and analysis of the demographic bias.

Considering the reconstruction task's evaluation, data sets containing 3D facial scans acquired in controlled conditions may be of some use. An example is the \textbf{FIDENTIS} data set \cite{FIDENTIS}, which licensed version provides textured one or even multiple 3D scans of 83 males and 117 females. In particular, it contains both raw frontal, profile, and merged models, the latter with and without ears. Furthermore, the models with the ears are provided with 42 associated landmarks, making them suitable for studies on photoanthropometry and the estimation of such landmarks (\textit{e.g.}, \cite{FIDENTISLandmarks}). Moreover, it is also suitable for the analysis of multi-session differences. However, a system evaluated on this data set could suffer from biased performance since, despite ages ranging between 18 and 67, 75\% of the subjects are aged between 21 and 29.

In order to evaluate the reconstruction methods under variations in lighting, occlusions, facial expression, acquisition environment, and viewing angle, it is possible to employ the \textbf{NoW benchmark} \cite{RingNet:CVPR:2019}. This data set contains 2054 2D images of 100 subjects (45 males and 55 females), captured with an iPhone X, and a 3D head scan for each subject as ground truth, captured through an active stereo system with the individual in a neutral expression. However, further demographic information about the subjects is not provided.

A data set which allows evaluating the reconstruction from videos is the \textbf{Florence 2D/3D} \cite{bagdanov2011florence}, providing 3D scans of 53 subjects (39 males and 14 females) and indoor and outdoor videos acquired in controlled and uncontrolled settings.
In particular, it is composed of four 3D models for each subject (\textit{i.e.}, two frontal, a right-side, and a left-side) and a further model with glasses whether he/she wears them. The HD videos (1280x720) were acquired indoors, at 25 FPS and four levels of zoom, while asking the subject to generate specific head rotations. The uncontrolled videos were acquired indoors at 25 FPS (704x576) and outdoors at 5-7 FPS (736x544), both at three levels of zoom and with the subject asked to be spontaneous. Hence, this data set could be employed in studies on reconstruction and recognition in realistic surveillance conditions, still lacking the occlusions which are typical of such a scenario. Moreover, the demographic bias could represent an issue since all the subjects are Caucasian and mostly aged between 20 and 30.

All the data sets described in this subsection consider one or more of the issues addressed in realistic forensic scenarios, both in terms of the environment (light conditions, indoor/outdoor), the subject (presence of occlusions, adverse poses, facial expressions), and technological factors (lower probe resolution, motion-blur, out-of-focus). Some of them also provide annotations related to forensic features (eye coordinates, facial parts), which could be useful in actual law courts \cite{ENFSI, FISWGmorph}. What is currently missing in the state-of-the-art data sets is the presence of occlusions of the lower face, such as in the case of facial masks, which could aid the research on the robustness to non-facial occlusions. 

\vspace{-0.4cm}
\section{Discussions}\label{sec:conclusion}

In this paper, we reviewed the state of the art of 3D face reconstruction (3DFR) from 2D images and videos for forensic recognition, evaluating the proposed approaches with respect to the requirements of a potential forensics-related system. Furthermore, the proposed approaches for enhancing forensic recognition in terms of performance were analyzed together with their potential application scenarios (Fig. \ref{fig:apptax}). 

The previously described studies mainly focus on enhancing the performance of recognition tasks in different contexts, such as the identification or verification of suspects within a gallery of mugshot images or the search for missing persons. They revealed the potential advantages of the fusion of the reconstructed model and the original images, which would allow taking advantage of the characteristics of a 3D facial model while limiting the possible loss of information in the reconstruction \cite{Liang2020}.
So far, researchers have proposed employing 3DFR either on the reference data or the probe material by re-projecting the 3D model into 2D images to aid a 2D-to-2D recognition. In particular, the first approach could find application in the adaptation of the pose of the model to the face in the probe material for easing both the visual comparison for investigative purposes and for employing the so-obtained figure in comparing it with the probe face through an automatic system, as preliminarily proposed for its application in forensic scenarios by Dutta et al. \cite{Dutta2012} and then further investigated by Zeng et al. \cite{Zeng2016, Zeng2017}. Similarly, this projection of reference 3D faces in the 2D domain in various poses demonstrated to improve the recognition performance of such systems, especially concerning their robustness to pose variation, introducing it as an augmentation step for training feature-based \cite{Zhang2008, HanJain2012} and DL systems \cite{Liang2018, Liang2020}. Moreover, Loohuis \cite{Loohuis2021} suggested that 3DFR could be successfully employed for mimicking the degraded quality of the probe data when coupled with rendering techniques for simulating such adverse conditions. The 3DFR from probe material finds applications in many scenarios as well, easing the comparison from a single probe image by rendering the face in order to match the pose with a reference image \cite{HanJain2012, Rahman2016} or by reconstructing it from multiple frames of a surveillance video \cite{vanDam2012, vanDam2016}.

Despite the promising results, especially concerning the robustness to pose variations in various probe and reference data types, most of the previously described studies did not evaluate their methods considering other requirements of an automated system supporting forensic analysis (Fig. \ref{fig:workseval}) related to understandability and forensic evaluation \cite{JainSurvey, forensicRequirements}, as summarized in Fig. \ref{fig:admissibility}. Moreover, the proposed methods do not assess their robustness to some typical issues of forensic cases, such as the presence of occlusions \cite{realocclusion, challenges}, making them inherently unsuitable for recognition scenarios involving them (Subsection \ref{subsec:admissibility}). However, some of them implicitly used a face recognition algorithm based on local descriptors \cite{Zhang2008, HanJain2012, Zeng2016, Zeng2017}, which supports the understandability of the output \cite{localmarks, interpretable}. Furthermore, a single study \cite{vanDam2016} employed a framework for easing forensic evaluation.

Although most of the proposed methods aim to enhance face recognition performance, they are not comparable quantitatively due to the variability in the considered settings. One of the most relevant differences is related to the involved data sets, which differ in acquisition environment, size, and availability (Section \ref{sec:databases}). 
The differences in terms of data type and quality represent another factor that makes them suitable for different tasks. Thus, it is necessary to address and compare these data sets separately (Subsection \ref{subsec:admissibility}) in terms of the recognition approach (Fig. \ref{fig:rectax}) and application scenarios (Sections \ref{sec:mugshot} and \ref{sec:other}).
Of course, differences are due to the time of publication, but recently the withdrawal of their availability due to more strict privacy rules on biometric data in the latter years made things complex. For example, the General Data Protection Regulation (GDPR) rules in the European Union strongly differ from those of other countries \cite{biosample,faceGDPR}. In particular, future studies should be based on data sets suitable for forensic research. The model, in our opinion, is the ForenFace data set \cite{ForenFace} because it takes realistic circumstances into account and also provides a set of anthropomorphic features proposed by the FISWG \cite{FISWGmorph}. Furthermore, they should evaluate the face reconstruction accuracy on large-scale 3D face data sets, such as the FIDENTIS one \cite{FIDENTIS}. 
Some forensic use cases are not yet included in any benchmark data set; for example, the special case of CCTV-based recognition from images recorded at ATMs with a very short distance from the subject and a distorted perspective \cite{automatedFaces}.

For both reconstruction and recognition tasks, a demographic analysis should be conducted on the performance to assess the bias against some demographic groups, an undesired issue in forensics that is sometimes overlooked even in current research \cite{JainSurvey}. To this aim, explicit demographic information about subjects represented 
in the data sets could aid in facing such an issue \cite{Scheuermann}. However, this useful data may be difficult to be assembled and recovered due to the privacy rules mentioned above. Moreover, the source of this bias could be related to the unbalancedness of the underlying data. This issue could be relieved by employing synthetic data sets, like the FAIR benchmark \cite{feng2022towards}. However, the employment of synthetic data still requires more investigation to be fully validated \cite{colbois2021use} and then accepted in the forensic context.

The eventual underlying 3D reference model could be affected by bias problems as well \cite{JainSurvey}, which may affect the face recognition system, making it unsuitable in forensic cases \cite{vanDam2016}.  
Therefore, a \textbf{model-free reconstruction} approach should be employed whenever possible. An example of this reconstruction approach is stereophotogrammetry, which allows capturing craniofacial morphology in high quality \cite{3Dstereo} to a level of detail that is often less important in generic recognition applications but which becomes crucial in the forensic context. Although it could not be suitable for its involvement in 3D-to-3D recognition scenarios, especially when based on shape comparison, such a reconstruction approach could be exploited in the generation of synthetic views for the comparison with the reference material \cite{vanDam2016} and, therefore, employed in a 2D-to-3D scenario. However, a drawback is the requirement of multiple images of the suspects \cite{vanDam2016}, which cannot be acquired in any forensic case. Another disadvantage of a model-free reconstruction approach is the significantly higher computational time required, making it unaffordable for real-time applications. Nonetheless, this represents a minor issue for many forensic applications, such as the ones related to lawsuits.

Thus, when a photometric reconstruction approach is unsuitable, a choice between approaches based on 3DMM and DL must be made \cite{uncalibrated}, even those not strictly proposed for forensic applications, taking into account their suitability, advantages and drawbacks.
For example, the methods based on 3DMM allow generating an arbitrary number of facial expressions, while those based on DL provide high-quality face texture synthesis \cite{uncalibrated, Geng2020}.
Therefore, the morphological model could be employed either for adapting the expression of the reference model or for imposing a neutral expression on the normalised face in the probe image, while a detailed reconstruction could be obtained through a DL network \cite{Geng2020}. However, it must be pointed out that huge manipulation, such as expression modification, could not be allowed in most evaluation cases, still being a valid aid for investigation purposes. These approaches have technical limits, namely the focus on global characteristics rather than fine details of the morphological model and the requirement of a great number of 3D scans for the training of DL networks \cite{uncalibrated}. The lack of understandability is another issue for the DL approach as well \cite{forensicRequirements}.
However, combining two or more reconstruction approaches could help limit some of the drawbacks of the single approach. For example, previous studies highlighted that it could be possible to reconstruct 3D faces that are highly detailed even with a single image by combining the prior knowledge of the global facial shape encoded in the 3DMM and refining it through a photometric approach \cite{Cao2018, Li2018, Rotger2019}. Similarly, the combination between a morphological model and one or more DL networks has been proposed as well \cite{Fan2021}. State-of-the-art methods not explicitly proposed for forensic applications should be further investigated in terms of potentialities and suitability as well, especially those based on DL, which revealed to be promising in addressing some of the typical issues in forensics like occlusion removal (\textit{e.g.}, \cite{Sharma20213d, mohaghegh2023robust}), 3DFR from one or multiple in-the-wild images (\textit{e.g.}, \cite{lin2020towards, zhang2021learning}), and face frontalization (\textit{e.g.}, \cite{yin2017towards}), thus potentially representing an aid in many investigative scenarios.

The computational time represents one of the main reasons why automated systems should be employed in forensics. It is an important feature in some specific applications, such as real-time identification through surveillance cameras (\textit{e.g.}, \cite{TheGuardianBeer}). In this regard, the online computational time must be assessed, representing the time required to test a single probe image and, thus, to recognize the captured individual. Specifically, it depends on both the recognition algorithm and the eventual strategy that must be applied to the probe to enhance the recognition task (\textit{e.g.}, some "canonical" representation).
In these terms, a reasonable computational time for some applications related to surveillance and lawsuit was reported by Zhang et al. \cite{Zhang2008} and Zeng et al. \cite{Zeng2016} (Table \ref{tab:mugtable}).
Zhang et al. \cite{Zhang2008} and Liang et al. \cite{Liang2018, Liang2020} also evaluated the offline computational time, representing the time required for applying the proposed enhancement approach based on 3D face reconstruction (\textit{e.g.}, gallery enlargement) and for the training of the recognition system. In particular, reported values suggest a notable improvement with respect to earlier proposals. However, despite these representing the most time-consuming processes, the offline computational time is generally of less concern since it does not impact real-time operations. 

It is important to remark that the most important feature in forensics is generally the reconstruction accuracy \cite{faceforensics, Ferkova2020, Suman2008}  since it represents a requirement which is often more strict than in generic recognition tasks.
In the literature, 3D model quality is evaluated from the errors in terms of shape by estimating the distance between the model and the corresponding ground truth. However, the extracted texture's quality should be assessed as well due to its role in the recognition task \cite{Afzal2020, 3Dstereo, ICPrecognition}. For example, the texture could allow exploiting facial marks, such as scars and tattoos. Their exploitation would enhance both performance and understandability in forensic comparison \cite{faceforensics, localmarks, marksEvidence, softMarks, FISWGmorph}. Furthermore, these facial marks are becoming even more valuable thanks to the availability of higher resolution sensors and the growing size of face image databases and their capability to improve speed and performance of recognition systems \cite{marksSurvey}. Hence, future research should take into account these additional features to assess their permanence in the generated 3D models.
This also holds for other morphological features, which forensic examiners evaluate to justify the outcome of the facial comparison (\textit{e.g.}, the decision whether the suspect is likely to be the one represented in a probe image) \cite{FISWGweb}. In addition to holistic ones (\textit{e.g.}, the overall shape), local characteristics are related to the proportions and the position of facial features, such as the relative size of the ears with respect to the eyes, nose, and mouth \cite{FISWGmorph}. The asymmetry between facial components should also be considered \cite{FISWGmorph}, thanks to its higher physical stability over time than other features. For example, the overall shape of the face could change because of the weight increase \cite{FISWGstability}; however, the asymmetry between facial components is less affected. Therefore, these features could be an effective aid for forensic examiners even to justify their conclusion on the comparison in law courts.

To sum up, we expect that great attention will be paid to the improvement of the recognition capability in forensic scenarios by 3DFR. Extremely unfavourable conditions, typically encountered in criminal cases, could be more affordable by considering both shape and texture appropriately modelled.
To this goal, data representative of forensic trace and reference material are necessary, also considering the robustness to other common factors altering the appearance, such as facial hair and the presence of occlusions. The bias toward a demographic group would be avoided in the data sets, favouring the system's fairness.
In our opinion, the proposed algorithms' understanding would couple with data availability. Data and algorithms will play a central role in effectively integrating 3D face reconstruction from 2D images and videos in the forensic field. Similarly, the employment of frameworks for easing forensic evaluation by non-expert professionals should become a practice for stressing the admissibility of the proposed methods in real cases. To this aim, an interdisciplinary approach involving computer science and law experts would speed up this process.
Therefore, we believe that its future involvement in real-world forensic applications is not far and that this survey contributes as a step toward this scenario.

\vspace{-0.3 cm}
\section*{Acknowledgements}

This work has been partially supported by the Italian Ministry of University and Research (MUR) within the PRIN2017-BullyBuster—A framework for bullying and cyberbullying action detection by computer vision and artificial intelligence methods and algorithms (CUP: F74I19000370001), and by SERICS (PE00000014) under the MUR National Recovery and Resilience Plan funded by the European Union - NextGenerationEU.
\vspace{-0.3 cm}

\bibliographystyle{ACM-Reference-Format}
\bibliography{sample-base}

\end{document}